\documentclass{article}

\usepackage[final]{neurips_2023}

\usepackage[utf8]{inputenc} 
\usepackage[T1]{fontenc}    
\usepackage{hyperref}       
\usepackage{url}            
\usepackage{booktabs}       
\usepackage{amsfonts}       
\usepackage{nicefrac}       
\usepackage{microtype}      
\usepackage{xcolor}         
\usepackage{wrapfig}
\usepackage{hyperref}
\usepackage{url}
\usepackage{amsmath,amsfonts,bm}
\usepackage{listings}
\usepackage{graphicx}
\usepackage{amssymb}
\usepackage{listings}
\usepackage{color}
\usepackage{makecell}
\usepackage{colortbl}
\usepackage{xcolor}
\usepackage{amsthm}
\usepackage{float}
\usepackage{wrapfig}
\usepackage{paralist}
\usepackage[figuresright]{rotating}
\usepackage{wrapfig,lipsum,booktabs}
\usepackage{subcaption}
\usepackage[labelformat=simple]{subcaption}
\usepackage{graphicx}

 \usepackage[square, comma, sort&compress, numbers]{natbib}
 \usepackage{algorithm}
\usepackage{algpseudocode}
\usepackage{algpascal}
\usepackage[misc]{ifsym}

\def\etc{\emph{etc.}}
\def\ie{\emph{i.e., }}

\definecolor{dkgreen}{rgb}{0,0.6,0}
\definecolor{gray}{rgb}{0.5,0.5,0.5}
\definecolor{mauve}{rgb}{0.58,0,0.82}

\usepackage{multirow}

\title{Hierarchically Gated Recurrent Neural Network for Sequence Modeling}

\author{
{
$^1$Zhen Qin$^{\star}$,\quad  $^2$Songlin Yang$^{\star}$,\quad
$^1$Yiran Zhong$^\textrm{\Letter}$
}\\
$^1$OpenNLPLab, Shanghai Artificial Intelligence Laboratory, $^2$MIT CSAIL \\
\texttt{https://github.com/OpenNLPLab/HGRN} 
}

\def\net{\textbf{HGRN}}
\def\name{\textbf{HGRU}}

\newcommand\blfootnote[1]{%
  \begingroup
  \renewcommand\thefootnote{}\footnote{#1}%
  \addtocounter{footnote}{-1}%
  \endgroup
}

\def\etc{\emph{etc.}}
\def\ie{\emph{i.e., }}

\begin{document}

\maketitle

\begin{abstract}

Transformers have surpassed RNNs in popularity due to their superior abilities in parallel training and long-term dependency modeling.
Recently, there has been a renewed interest in using linear RNNs for efficient sequence modeling.
These linear RNNs often employ gating mechanisms in the output of the linear recurrence layer while ignoring the significance of using forget gates within the recurrence. In this paper, we propose a gated linear RNN model dubbed Hierarchically Gated Recurrent Neural Network (HGRN), which includes forget gates that are lower bounded by a learnable value. The lower bound increases monotonically when moving up layers. This allows the upper layers to model long-term dependencies and the lower layers to model more local, short-term dependencies. Experiments on language modeling, image classification, and long-range arena benchmarks showcase the efficiency and effectiveness of our proposed model. The source code is available at \href{https://github.com/OpenNLPLab/HGRN}{https://github.com/OpenNLPLab/HGRN}\blfootnote{\noindent $^{\star}$Equal contribution. $^\textrm{\Letter}$ Indicates corresponding author (Email address: \textit{zhongyiran@gmail.com}).}.

\end{abstract}

\section{Introduction}
Sequence modeling is a fundamental problem in various domains such as natural language processing~\cite{devlin-etal-2019-bert,liu2019roberta,qin2023scaling,qin-etal-2023-linear,liu2022neural}, time series analysis~\cite{zhou2021informer}, computer vision~\cite{vit,vivit,Sun2023Tpami,lu2022linear}, and audio processing~\cite{gong21b_interspeech,akbari2021vatt, sun2022locality}. Prior to the invention of Transformers~\cite{vaswani2017attention}, RNN and its variants were the primary selections of architectures for sequence modeling, and have been widely used in machine translation~\cite{cho-etal-2014-learning}, stock price prediction~\cite{selvin2017stock}, weather forecasting~\cite{salman2015weather}, speech recognition~\cite{miao2015eesen}, and \etc 

RNNs have two main drawbacks: slow sequential training and limited capability in modeling long-term dependencies.  With the swift development of deep learning and the pervasive use of GPUs, these drawbacks prevent it from flourishing in modern long-sequence modeling tasks. Meanwhile, Transformers \cite{vaswani2017attention} have rapidly gained popularity and now dominate various research areas in sequence modeling due to their better abilities in parallel training and long-term dependency modeling.
However, Transformer's quadratic time complexity makes long sequence modeling expensive. On the other hand, RNN offers linear complexity and serves as an ideal choice for long sequence modeling. 
This work aims to address these RNN drawbacks,  revitalizing their applicability in long-sequence modeling tasks.

To address the training inefficiency problem, we turn to more efficient RNN variants that employ element-wise linear recurrence (ELR) relations \cite{iclr18}. ELR provides two main advantages: (i)
Removing nonlinearities in the recurrence enables parallelized training. (ii)
By assuming independence between distinct hidden states, it enables 
 efficient hidden state updates (through element-wise product instead of matrix multiplication) \cite{lei-etal-2018-simple, s4d}.
Notably, ELR has been used in many modern linear RNN models, including the diagonalized versions of structured state-space models \cite{s4} (S4) \cite{dss, s4d, s5} and RWKV \cite{rwkv}.
In recent advancements, numerous studies have incorporated gating mechanisms into the outputs of linear recurrence layers \cite{gss, h3, mega, pretrainingwoattn, rwkv}, similar to the output gates in LSTMs and leading to considerable performance gains. However, most current studies overlook the significance of the forget gate, which is often regarded as the most important gate in LSTMs \cite{Greff2015LSTMAS, forgetgate}. In this work,
we underscore the importance of employing forget gates in linear RNNs and adopt gated linear RNNs for both efficiency and high performance.

To effectively capture long-term dependencies in gated RNNs, it is crucial to maintain high forget gate values close to one \cite{gu-improving}. However, gates in saturated regimes (i.e., close to zero or one) suffer from the gradient vanishing issue \cite{gu-improving}. Moreover, if all forget gate values are close to one, RNNs will not be able to effectively forget irrelevant information, compromising their ability to model short-term dependencies.
To address these challenges, we introduce Hierarchically Gated Recurrent Units(\name). In \name, we add an additive learnable value, referred to as the lower bound, to the original forget gate value, effectively mitigating the issue of saturated gates \cite{gu-improving} by pushing gate activations away from the saturated regimes.
Furthermore, we design the lower bounds to increase monotonically as we move up the layers of the RNN. This ensures that the forget gate values in the lower layers remain relatively small, enabling the necessary forgetting of past information for modeling short-term dependencies. In contrast, in the uppermost layer, the forget gate values approach one, facilitating the effective modeling of long-term dependencies. 
Our proposed model has proven to be highly efficient and effective, as demonstrated by its outstanding performance in language modeling, image classification, and long-range arena benchmarks.

\section{Related work}

\paragraph{Efficient token mixing for sequence modeling.} 
\cite{metaformer} abstracts self-attention (SA) as token mixing, thereby transforming the Transformer architecture into MetaFormer. MetaFormer comprises essential components such as token mixer, channel mixer, residual connections, and LayerNorm. This abstraction highlights that the success of Transformers does not solely rely on SA but rather on the holistic integration of these components. Notably, token mixers can be replaced with simpler alternatives like pooling layers without compromising the model's performance in the context of vision transformer. For sequence modeling tasks, \cite{unified} provides a comprehensive analysis and discussion of different token mixing strategies. Two prominent contenders, long convolution and linear recurrence, show promise as replacements for SA modules in long sequence modeling due to their superior asymptotic time complexity and competitive performances. In long convolution models \cite{Li2022WhatMC, simplelongconv, qin2023toeplitz, hyena}, the kernel size matches the input sequence length, enabling a broader context compared to traditional convolutions. Training is accomplished using the efficient $\mathcal{O}(n\log n)$ fast Fourier transforms (FFT) algorithm. However, long convolutions face challenges such as the need for causal convolution inference, which requires caching all historical computations similar to the key-value (KV) cache in SA. This can lead to memory limitations when processing long sequences. Moreover, the inference complexity of long convolutions remains higher than that of RNNs. These factors make linear RNNs a more suitable alternative to replace SA in long-sequence modeling. TransNormerLLM~\cite{qin2023scaling} scales efficient token mixing in large language models to achieve competitive performance and superior training and inference efficiency compared to transformer-based models.

\paragraph{Element-wise linear recurrence.} The slower training speeds of traditional RNNs can be attributed to two main factors: (i) The updating of the hidden state involves full matrix multiplication.
(ii) The presence of nonlinearity within the recurrence prevents parallel computation.
To tackle the first issue, \cite{lei-etal-2018-simple} introduced a simplified interaction between hidden states. This allowed the hidden state update to be performed using an element-wise product instead of full matrix multiplication. They demonstrated that this approach is notably fast when the (nonlinear) recurrence for each dimension is fused within a single CUDA kernel. Likewise, for the linear case, diagonalized versions of S4 \cite{dss, s4d} have also exhibited speed improvements over S4 by leveraging element-wise recurrence. Regarding the second challenge, the ability to capture nonlinear dependencies on past data can be achieved by stacking multiple linear recurrence layers interleaved with nonlinear MLP blocks. This indicates the potential to eliminate nonlinearity, as suggested by \cite{strongtypedrnn, iclr18, 2110.13985}. Empirical support for this strategy's effectiveness came later, as demonstrated by \cite{DBLP:conf/nips/GuJGSDRR21, s4d, s5, lru, h3, rwkv}.
\cite{DBLP:journals/corr/abs-2307-11888} further highlighted that such an architecture still possesses Universal Approximator properties, thus justifying the employment of linear recurrence. By excluding nonlinearity, \cite{iclr18, s5} showed that the parallel scan algorithm can be used for parallel training. 

Linear recurrence can be broadly categorized into exponential moving averages (EMA) and gating schemes, as noted by \cite{iclr18}. The key difference is whether the decay rate is data-dependent.
Models such as S4 \cite{s4}, S4D \cite{s4d}, MEGA \cite{mega}, RWKV \cite{rwkv}, and LRU \cite{lru} utilize the EMA approach, where the decay rate is static for all time steps (i.e., data-independent), while our model uses a data-dependent dynamic decay rate through the use of the forget gate. We remark on the importance of incorporating a data-dependent decay rate, which is largely ignored by current works in linear RNNs. Although liquid S4 \cite{liquids4} uses a dynamic transition matrix (which amounts to a data-dependent decay rate), it employs a limited form for FFT-based training. Our model does not have the convolutional view and thus cannot use FFT for training but allows the use of parallel scans.

The field of linear Transformers and linear RNNs exhibits a close relationship. \cite{xfmrsarernns} shows that linear Transformers can be reformulated as RNNs during auto-regressive decoding, revealing similarities to the update rules observed in fast weight additive outer products \cite{Schmidhuber1992LearningTC, Schlag2021LinearTA}. These updates can be seen as a special case of element-wise linear recurrence, where forget gate values are consistently set to one across time and hidden states are two-dimensional. However, this formulation in linear Transformers lacks the ability to forget irrelevant information, resulting in the attention dilution issue \cite{qin-etal-2022-devil}. To address this limitation, \cite{Schlag2021LinearTA} introduced the delta rule to forget values associated with the current write key by removing the corresponding value before adding the new value. Alternatively, \cite{rfa, mao-2022-fine} proposed gating mechanisms similar to those in gated RNNs to facilitate the forgetting of irrelevant information. 

\paragraph{Long-term dependencies in RNNs.} 
RNNs fall short in long-term dependency modeling, which is commonly attributed to the gradient vanishing issue. Three methods are typically applied to mitigate this issue. (i) Gating mechanisms \cite{DBLP:journals/neco/GersSC00, Hochreiter2001GradientFI, gru, onlstm, gu-improving}, which are believed to be crucial to the success of LSTMs, use additive (instead of multiplicative) hidden state update rules to improve gradient flow.
(ii) Regularizing or initializing the eigenvalues of the recurrent weight matrix (close) to one via identity matrices \cite{Le2015ASW} or unitary matrices \cite{DBLP:conf/icml/ArjovskySB16}. In the diagonal linear RNN case, the eigenvalues coincide with the element-wise decay rates, and LRU \cite{lru} uses randomized linear algebra techniques to initialize eigenvalues to be close to one. \cite{lru} also interestingly points out that many modern state-space models use a very small time step value on initialization for discretization, resulting in eigenvalues or decay rates close to one. 
(iii) Adding skip connections between distant time steps to allow shortcuts for gradient flow \cite{DBLP:conf/icml/KoutnikGGS14,dilatedrnn,DBLP:conf/iclr/ChungAB17}.Our approach combines (i) and (ii), which improves gating mechanisms with a regularized dynamic decay rate that approaches one in the upper layer.

\section{Method}
\subsection{Architecture overview} 
\begin{figure}[h]
    \centering
    \includegraphics[width=\textwidth]{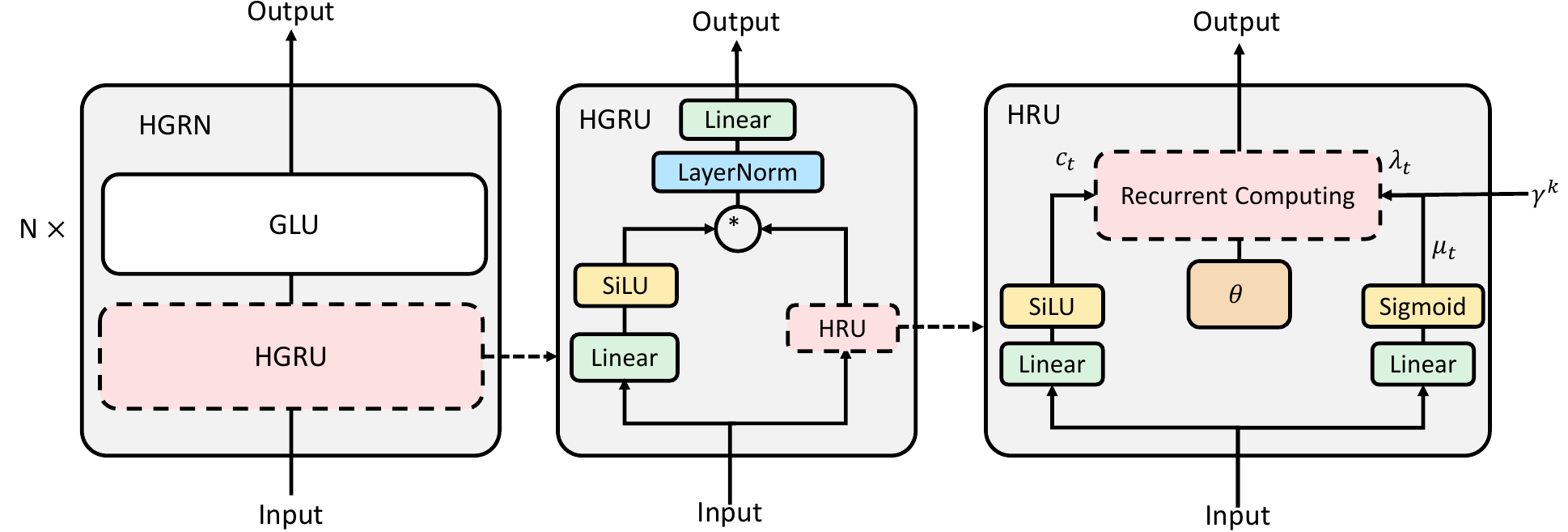}
    \caption{Illustration of the neural architecture.
    Each {\net} layer consists of a token mixer {\name} and a channel mixer GLU.     {\name} employs linear recurrence in the complex domain: $  \mathbf h_t  = \lambda_t \odot \exp(i\theta) 
\odot \mathbf h_{t-1}
+(1-\lambda_t) \odot \mathbf  c_t$.  Here $c_t$ is the input vector, $\theta$ is the rotation angle, $\mu_t$ is the output of the original forget gate, $\gamma^k$ is the lower bound of the $k$-th layer, $\lambda$ is the resulting data dependent decay rate: $\lambda_t = \gamma^k + (1 - \gamma^k) \odot \mu_t$. 
    }
    \label{fig: network}
\end{figure}

Our proposed Hierarchically Gated Recurrent Network ({\net}) is depicted in Figure~\ref{fig: network}. It has multiple stacked layers, each of which consists of a token mixing module {\name} and a channel mixing module \textbf{GLU} (Gated Linear Unit ~\cite{glu}).

\begin{wrapfigure}[9]{r}{0.4\textwidth}
\begin{minipage}{0.4\textwidth}
\vspace{-8mm}
\begin{algorithm}[H]
\caption{Recurrent Computing}
\begin{algorithmic}[1]
\State
Input: $\mathbf c_t \in \mathbb C^{1\times d}, \mu_t , 
\theta, \gamma^k \in \mathbb R^{1\times d}, t=1,\ldots, n, k=1,\ldots,H .$ \\
Init: $\mathbf h=\mathbf 0 \in \mathbb C^{1\times d}, \mathbf H \in \mathbb C^{n\times d}$.
\For{t = 1}{n}
\Begin 
\State  $\lambda_t = \gamma^k + (1 - \gamma^k) \odot \mu_t$.
    \State $\mathbf h=\lambda_t \exp(i\theta)\mathbf h + (1-\lambda_t) \mathbf c_t$. 
    \State $[\mathbf H]_t = \mathbf h$.
 \End \\
return $\mathbf H$.
\end{algorithmic}
\end{algorithm}
\end{minipage}
\label{algorithm:lr}
\end{wrapfigure}

\subsection{HGRU exploration}
We begin with a simple gated linear recurrent layer, which is defined as:
\begin{equation}
\begin{aligned}
 \mathbf f_t & =\mathrm{Sigmoid}\left( \mathbf x_t \mathbf W_f+\mathbf b_f\right)\in \mathbb R^{1\times d}, \\
\mathbf i_t & =\mathrm{Sigmoid} \left(\mathbf x_t \mathbf W_i +\mathbf b_i\right)\in \mathbb R^{1\times d}, \\
\mathbf c_t & =\mathrm{SiLU}\left(\mathbf  x_t \mathbf W_t + \mathbf b_z\right) \in \mathbb R^{1\times d},\\
\mathbf h_t & =\mathbf f_t \odot \mathbf h_{t-1}+ \mathbf i_t \odot \mathbf c_t\in \mathbb R^{1\times d},\\
\mathbf h_0&=\mathbf 0 \in \mathbb R^{1\times d},
\end{aligned}
\end{equation}
where $\odot$ denotes the element-wise product. 
Following the terminology used in the RNN literature, we refer to $\mathbf f_t$ and $\mathbf i_t$ as the forget and input gates, respectively. It is worth noting that $\mathbf f_t$ and $\mathbf i_t$ depend only on $\mathbf x_t$ and not on $\mathbf h_{t-1}$. This characteristic enables the use of the parallel scan algorithm \cite{iclr18, s5}, otherwise it is infeasible. We then make the following changes toward our final {\name} step by step.

\paragraph{Complex-valued recurrence.} For linear RNNs with static decay rates, it is common to perform eigendecompositions on the recurrent weight matrix to achieve element-wise linear recurrence. However, if only real-valued eigenvalues are allowed, it restricts the range of the recurrent weight matrix to be symmetric, limiting the model's expressiveness. To overcome this limitation, linear RNNs often employ complex-valued eigenvalues to enhance the model's expressive power \cite{s4d,dss,lru,huang2023encoding,DBLP:journals/corr/abs-2212-00768}. Motivated by this, we extend our model to consider $\mathbf{h}_t, \mathbf{i}_t, \mathbf{c}_t \in \mathbb{C}^{1\times d}$ as complex values.
 For the input $c_t$, we parameterize its real and imaginary parts separately as follows:
\begin{align*}
      \mathrm{Re}(\mathbf{c}_t)&= \mathrm{SiLU}\left( \mathbf x_t \mathbf W_{cr} +\mathbf b_{cr}\right) \in \mathbb R^{1\times d}, \\
    \mathrm{Im}(\mathbf{c}_t)&= \mathrm{SiLU}\left( \mathbf x_t \mathbf W_{ci}+\mathbf b_{ci}\right) \in \mathbb R^{1\times d}.
\end{align*}
Regarding the forget gate values, we find it convenient to use the exponential representation of complex numbers and parameterize $\mathbf f_t$ as follows:
$\mathbf f_t =  \lambda_t \odot \exp (i \theta_t)$. 
Here, $i^2=-1$, ${\lambda_t}, {\theta_t} \in \mathbb{R}^{d} $ and $\exp (i \theta_t) = \cos \theta_t + \sin \theta_t i$. The magnitude argument $\lambda_t$ determines the intensity of remembering historical information, while the phase argument $\theta_t$ determines the oscillation frequencies. We find that parameterizing $\theta_t$ in a data-independent manner is preferable, as it allows for a clear interpretation of encoding relative position information (see next subsection for more discussions)
, which is reminiscent of Rotary Positional Embedding (RoPE) \cite{rope}. We shared $\theta$ arcoss times steps, i.e., $\mathbf f_t =  \lambda_t \odot \exp (i \theta)$, initialize $\theta$ as RoPE does, but make it learnable like LRPE \cite{qin2023linearized}.

\paragraph{Lower bound on forget gate values.}
Since the intensity of remembering information is only related to the magnitude argument $\lambda_t$, we focus on how to add a lower bound to $\lambda_t$. As mentioned earlier, we want to set a monotonically increasing lower bound on the forget gate (magnitude) values. Inspired by ON-LSTM \cite{onlstm},  we employ the \texttt{cummax} activation function to achieve this.   Concretely, we allocate $\mathbf \Gamma \in \mathbb R^{H \times d}$ to parameterize lower bounds independently for all hidden states, where $H$ is the number of layer. Assuming the layer index is $k$, we have the following calculations:
\begin{align*}
\mathbf P &=(\mathrm{Softmax}(\mathbf \Gamma, \mathrm{dim=0}) \in \mathbb R^{H \times d}, \\
 \gamma^k &=[\mathrm{Cumsum}(\mathbf P, \mathrm{dim=0})]_k \in \mathbb R^{1\times d}.
\end{align*}
Here we define $[\mathrm{Cumsum}(\mathbf x)]_k=(\sum_{i=1}^{k} x_i) - x_1$ to prevent the highest layer's lower bound from being one as we still want the ability to forget irrelevant information. 

We remark that there is a difference in the use of \texttt{cummax} between our model and ON-LSTM. In ON-LSTM, \texttt{cummax} is applied to the hidden state dimension within a single layer, while in our case, we apply \texttt{cummax} on the layer dimension across different layers to enable upper layers to model long-range dependencies.  

Finally, $\lambda_{t}$ in the $k$-th layer is parameterized as follows:
\begin{align*}
     \mu_t &=\mathrm{Sigmoid}\left(\mathbf x_t \mathbf W_{\mu}+ \mathbf b_{\mu}\right) \in \mathbb R^{1\times d},\\
    \lambda_{t}&= \gamma^k + (1 - \gamma^k) \odot \mu_t  \in \mathbb R^{1\times d}.
\end{align*} 

Compared to before (i.e., without lower bounds), to achieve the same forget rate value $\bar \gamma$ closed to one, $\mu_t$ will be pushed away from the Sigmoid activation function's saturated regions (i.e., near one),
\[
\mu_t = \frac{\bar \gamma -\gamma^k}{1-\gamma^k}< \bar \gamma,
\]
thereby mitigating the gradient vanishing issue \cite{gu-improving} and making gradient-based optimization easier.

\paragraph{Tying input and forget gates.} To reduce the number of parameters, it is common to use leaky units, i.e., tying the input gate with the forget gate using $\mathbf i_t = 1- \mathbf f_t$, which has a close relationship to the discretization of continuous-time system \cite{DBLP:conf/iclr/TallecO18a} and exponential moving average \cite{Hunter1986TheEW}, and has been proven effective empirically \cite{gru, Greff2015LSTMAS}. To allows for a clear interpretation of encoding relative position information, we only apply this strategy on the magnitude argument: 
\begin{equation}
    \mathbf h_t  = \lambda_t \odot \exp(i\theta) 
\odot \mathbf h_{t-1}
+(1-\lambda_t) \odot \mathbf  c_t  \in \mathbb C^{1\times d}.
\label{eq-rnn}
\end{equation}

\paragraph{Output gates and projection.} The addition of gates to the output of the recurrence layer has been shown to be effective in state-space models \cite{gss, h3, mega, pretrainingwoattn}. Motivated by this, we incorporate an output gate before performing the output projection as follows and get $\name$:

\begin{equation}
\begin{aligned}
\mathbf g_t &= \mathrm{Sigmoid}( W_g \mathbf x_t + b_g)  \in \mathbb R^{1\times 2d},  \\
\mathbf o^{\prime}_t &=\mathrm{LayerNorm}(\mathbf  g_t \odot [\mathrm{Re}(\mathbf h_t), \mathrm{Im}(\mathbf h_t)]) \in \mathbb R^{1\times 2d}, \\
\mathbf o_t  &=   \mathbf o^{\prime}_t\mathbf W_{o} + \mathbf b_o \in \mathbb R^{1\times d}.
\end{aligned}
\end{equation}

\subsection{Token mixing perspective of {\name}}
We provide the token mixing perspective of {\name} similar to  
 \cite{huang2023encoding}. 
Expanding Equation \ref{eq-rnn}, we have:
\begin{equation}
\mathbf h_t  = \sum_{s=1}^t (1-\lambda_s)\left[ \prod_{k=s+1}^t\lambda_k\exp(i\theta) \right ] \mathbf  c_s =
 \sum_{s=1}^t (1-\lambda_s)\left[ \prod_{k=s+1}^t\lambda_k\right ]\exp(i (t-s)\theta)  \mathbf  c_s
\end{equation}
Written in matrix form, we have:
\begin{equation}
\begin{aligned}
\mathbf H=\left[
\begin{matrix}
\mathbf h_1 \\
\vdots\\
\vdots\\
\mathbf h_n
\end{matrix}
\right],
\mathbf A =\left[\begin{matrix}
1-\lambda_1 & 0 & \cdots  & 0   \\
(1-\lambda_1)\lambda_2 \exp(i \theta) &  1-\lambda_2 &  &  \vdots \\
\vdots &\vdots   &\ddots&  0 \\
(1-\lambda_1) \left[ \prod_{k=2}^n \lambda_k  \right] \exp(i(n-1)\theta)& 
\ldots  & 
\ldots  & 
1-\lambda_{n}
\end{matrix}\right], 
\mathbf C=\left[
\begin{matrix}
\mathbf c_1 \\
\vdots\\
\vdots\\
\mathbf c_n
\end{matrix}
\right]
\end{aligned}
\end{equation}
So the token mixing module can be formed as follows:
\begin{equation}
\mathbf H =\mathbf A \mathbf C.
\end{equation}
Note that the token mixing matrix $\mathbf A$ can be decomposed into two parts $\mathbf A = \mathbf \Lambda \odot \mathbf \Theta$:
\begin{equation}
\begin{aligned}
\mathbf \Lambda  =
\left[\begin{matrix}
1-\lambda_1 & 0 & \cdots  & 0   \\
(1-\lambda_1)\lambda_2  &  1-\lambda_2 &  &  \vdots \\
\vdots &\vdots   &\ddots&  0 \\
(1-\lambda_1) \left[ \prod_{k=2}^n \lambda_k  \right]& 
\ldots  & 
\ldots  & 
1-\lambda_{n}
\end{matrix}\right],
\mathbf \Theta =
\left[\begin{matrix}
1& 0 & \cdots  & 0   \\
\exp(i \theta) &  1 &  &  \vdots \\
\vdots &\vdots   &\ddots&  0 \\
\exp(i(n-1)\theta)& 
\ldots  & 
\ldots  & 
1
\end{matrix}\right]
\end{aligned}
\end{equation}
This decomposition means that the Token mixing matrix $\mathbf \Lambda$  can be decoupled into two independent
modules, where $\mathbf \Lambda$ models the long-distance dependency and $\mathbf \Theta$, a Toeplitz matrix, models the relative positional relationship and enhanced expressiveness. Note that if $\mathbf \Theta$ depends on the input, then the matrix $\mathbf \Lambda$ will no longer be a Toeplitz matrix, thus unable to capture relative position information. It can be also viewed as a RoPE-enhanced attention mechanism: $\mathbf \Lambda$ corresponds to the attention matrix but the attention score here is the cumulative product of data-dependent decay rates; $\mathbf \Theta$ directly corresponds to RoPE.

\section{Experiments}
We conduct a comparative analysis between our proposed {\net} and four widely adopted sequence modeling structures, \ie attention-based, MLP-based, FFT-based, and state-space-based.
We evaluate {\net} on the WikiText-103 dataset \cite{merity2017pointer} and the Pile~\citep{pile} dataset for autoregressive language modeling, as well as the length extrapolation ability.
To assess the accuracy and efficiency of our model in handling long-term dependencies, we utilize the LRA benchmark~\cite{lra}.
Additionally, we showcase the robustness of {\net} in computer vision task on the ImageNet-1k dataset. 

\subsection{Setting}
We implement our models in Pytorch~\cite{paszke2019pytorch} and train them on 8 Nvidia A100 GPUs. For \net, we found that fusing element-wise recurrence into a single CUDA kernel results in fast running speed in practice. \cite{iclr18} also found that unless the sequence length is sufficiently large, the parallel scan's implementation is not necessarily faster than the sequential scan. As such, we use a CUDA-based sequential scan for implementation; however, our model has the potential to model very long sequences through the use of a parallel scan.

We adopt the same training configuration for all competitors, including batch size, learning rate, training epochs or iterations, \etc~ We list detailed hyper-parameters in the Appendix. 
For the autoregressive language modeling, we conducted three sets of experiments. Firstly, we validated the performance of two different-scale models on the Wikitext-103 dataset. We used the TNN configuration to verify the performance of the model at around 44m, and the Hyena configuration to verify the performance of the model at around 125m. To evaluate the performance of larger-scale models, we trained a 1b Transformer and {\net} on the Pile dataset using 10b tokens. To assess the performance in downstream tasks, we trained {\net}  models of 150m, 350m, and 1b on the Pile dataset using 100b tokens and conducted zero-shot evaluations on downstream tasks.

For the LRA benchmark, 
We report results on all 6 tasks.
For the image classification on the ImageNet-1k dataset, 
We integrate {\net} into the DeiT~\cite{pmlr-v139-touvron21a} structure, we replace the transformer layers with our {\net} modules.
It is compared
to the performance of the vanilla DeiT
on the ImageNet-1K dataset for image classification.

\subsection{Results}
\label{results}
\paragraph{Autoregressive Language Modeling}

\begin{table}[t]
    \centering
    \small  
    \caption{\textbf{Results on Wikitext-103} (TNN\cite{qin2023toeplitz}'s setting). $\downarrow$ means \textit{lower is better}.}
     \setlength{\tabcolsep}{1.1cm}
     \label{lm}
    \begin{tabular}{l|lll}
    
    \hline
        Model & \makecell[c]{PPL \\(val)$\downarrow$} & \makecell[c]{PPL \\(test)$\downarrow$} &  \makecell[c]{Params\\(M)} \\ \hline
        \textit{Attn-based}  \\ \hline
        Transformer~\cite{vaswani2017attention} & 24.40 & {24.78} & 44.65 \\ 
        FLASH~\cite{dao2022flashattention} & 25.92 & 26.70 & 42.17 \\ 
        1+elu~\cite{katharopoulos2020transformers} & 27.44 & 28.05 & 44.65 \\ 
        Performer~\cite{choromanski2020rethinking} & 62.50 & 63.16 & 44.65 \\
        cosFormer~\cite{zhen2022cosformer} & 26.53 & 27.06 & 44.65 \\ \hline
        \textit{MLP-based}  \\ \hline
        Syn(D)~\cite{tay2021synthesizer} & 31.31 & 32.43 & 46.75 \\ 
        Syn(R)~\cite{tay2021synthesizer} & 33.68 & 34.78 & 44.65 \\ 
        gMLP\cite{liu2021pay} & 28.08 & 29.13 & 47.83 \\ \hline
        \textit{RNN-based}  \\ \hline
        S4~\cite{gu2022efficiently} & 38.34 & 39.66 & 45.69 \\ 
        DSS~\cite{dss} & 39.39 & 41.07 & 45.73 \\ 
        GSS~\cite{gss} & 29.61 & 30.74 & 43.84 \\ 
        RWKV~\cite{rwkv} & 24.31 & 25.07 & 46.23\\
        LRU~\cite{lru} & 29.86 & 31.12 & 46.24\\
        \hline
       \textit{FFT-based}  \\ \hline
       TNN~\cite{qin2023toeplitz} & {23.98} & {24.67} & 48.68 \\ \hline
        \textit{Ours} & ~ & ~ & ~ \\ \hline
        {\net} & {24.14} & 24.82 &46.25 \\ \hline
    \end{tabular}
\end{table}

\begin{wraptable}[14]{r}
{0.5\textwidth}
    \caption{\textbf{Results on Wikitext-103} (Hyena\cite{hyena}'s setting). All models are in GPT-2 small size (125M). $\downarrow$ means \textit{lower is better}} 
    \centering
    \label{lm:hyena}
     \setlength{\tabcolsep}{0.8cm}
     \vspace{-2mm}
    \begin{tabular}{l|l l }
    \hline
        Model & PPL$\downarrow$ \\ \hline
        Transformer & {18.6} \\ 
        Hybrid H3 & {18.5}  \\ 
        Performer & 26.8  \\ 
        Reformer & 25.6  \\ 
        AFT-conv & 28.2  \\ 
        Linear Attention & 25.6 \\ 
        Hyena & {18.6} \\
        Hyena-slim & {18.5}  \\ \hline
        HGRN &{18.6} \\ 
        \hline
    \end{tabular}
\end{wraptable}

Autoregressive language modeling stands as a prominent task within the field of natural language processing, as it serves as a measure of a language model's causal inference capability.
This task requires the model to estimate the probability distribution of the subsequent token based on the previously seen tokens.

We show the performances of the autoregressive language modeling in table~\ref{lm} and table~\ref{lm:hyena}.
Compared to transformer-based methods, {\net} performs favourably than most efficient variants of the vanilla transformer such as FLASH~\cite{hua2022transformer}, 1+elu~\cite{katharopoulos2020transformers}, Performer~\cite{choromanski2020rethinking} and cosFormer~\cite{zhen2022cosformer}. 
Also, {\net} achieves better results than the MLP-based methods with a notable margin.
Nevertheless,
{\net} performs similarly to the original transformer~\cite{vaswani2017attention}.
Finally, {\net} shares similar concepts with RNN-based such as S4~\cite{gu2022efficiently}, DSS~\cite{dss}, GSS~\cite{gss}, RWKV~\cite{rwkv}, and LRU~\cite{lru}, our {\net} also achieves superior performance to all RNN-based methods. This provides evidence HRGN may be an effective method in LM
We also report the extrapolation ability of {{\net}} compared to previous methods in Table~\ref{extrapola}.

\begin{wraptable}[7]{hbt!}
{0.5\textwidth}
\label{lm:pile}
\small
\vspace{-4mm}
    \caption{\textbf{Results on the Pile.} All the model size is 1b. The lower the better.}
    \vspace{-2mm}
    \centering
    \setlength{\tabcolsep}{1.2cm}
    \begin{tabular}{l|ll}
    \hline
         Model & PPL$\downarrow$ \\ \hline
        Transformer & 4.56  \\
        LRU & 5.07  \\
        \hline
        \net & {4.14}  \\  \hline
    \end{tabular}
\end{wraptable}

We also trained a 1b model on the Pile dataset and compared it with LRU and Transformer. Specifically, our training parameters included a sequence length of 1024, batch size of 96, 100k updates, and a learning rate of 5e-4. It can be seen that {\net} still performs better at the 1b scale. Additionally, we trained 100b tokens of {\net} on the Pile dataset at 150m, 350m, and 1b sizes, and evaluated them against open-source Transformer-based models in downstream tasks. We selected Comparison on Commonsense Reasoning and Super GLUE tasks, and all evaluations were done using the lm-evaluation-harness~\cite{eval-harness}. {\net} achieves comparable performance to Transformer-based models when consuming only 1/3 of the tokens.

\begin{table}[t]
\label{lm:csr}
    \centering
    \small
    \caption{\textbf{Performance Comparison on Commonsense Reasoning.}. PS: parameter size (billion). T: tokens (billion).
    HS: HellaSwag. WG: WinoGrande. }
    \setlength{\tabcolsep}{0.2cm}
    \begin{tabular}{l|ll|llllllll}
    \hline
        Model & PS & T & BOOLQ & PIQA & HS & WG & ARC-e & ARC-c & OBQA & AVG \\ \hline
        GPT-Neo & 0.13 & 300 & 61.71  & 63.06  & 30.40  & 50.43  & 43.73  & 23.12  & 26.20  & 42.66  \\ 
        OPT & 0.16 & 300 & 55.47  & 62.95  & 31.35  & 50.43  & 43.52  & 22.70  & 28.00  & 42.06  \\ 
        Pythia & 0.16 & 300 & 55.08  & 61.32  & 30.16  & 51.93  & 43.18  & 23.12  & 26.80  & 41.66  \\ 
        RWKV & 0.17 & - & - & 65.07  & 32.26  & 50.83  & 47.47  & 24.15  & 29.60  & 41.56  \\ 
        HGRN & 0.15 & 100 & 59.91  & 65.02  & 33.33  & 50.20  & 46.68  & 23.81  & 28.60  & 43.94  \\ \hline
        OPT & 0.35 & 300 & 57.74  & 64.58  & 36.69  & 52.49  & 44.02  & 23.89  & 28.20  & 43.94  \\ 
        Pythia & 0.4 & 300 & 60.40  & 67.08  & 40.52  & 53.59  & 51.81  & 24.15  & 29.40  & 46.71  \\ 
        BLOOM & 0.56 & 350 & 55.14  & 64.09  & 36.97  & 52.80  & 47.35  & 23.98  & 28.20  & 44.08  \\ 
        RWKV & 0.43 & - & - & 67.52  & 40.90  & 51.14  & 52.86  & 25.17  & 32.40  & 45.00  \\ 
        HGRN & 0.35 & 100 & 59.05  & 66.70  & 38.12  & 51.70  & 49.20  & 25.26  & 30.60  & 45.80  \\ \hline
        GPT-Neo & 1.3 & 300 & 61.99  & 71.11  & 48.93  & 54.93  & 56.19  & 25.85  & 33.60  & 50.37  \\ 
        OPT & 1.3 & 300 & 57.77  & 71.71  & 53.70  & 59.35  & 57.24  & 29.69  & 33.20  & 51.81  \\ 
        Pythia & 1.4 & 300 & 60.73  & 70.67  & 47.18  & 53.51  & 56.99  & 26.88  & 31.40  & 49.62  \\ 
        BLOOM & 1.1 & 350 & 59.08  & 67.14  & 42.98  & 54.93  & 51.47  & 25.68  & 29.40  & 47.24  \\ 
        RWKV & 1.5 & - & - & 72.36  & 52.48  & 54.62  & 60.48  & 29.44  & 34.00  & 50.56  \\ 
        HGRN & 1 & 100 & 58.69  & 70.89  & 48.02  & 51.62  & 55.64  & 27.90  & 31.60  & 49.19 \\ \hline
    \end{tabular}
\end{table}

\begin{table}[!ht]
\label{lm:sg}
    \centering
    \caption{\textbf{Performance Comparison on SuperGLUE.} PS: parameter size (billion). T: tokens (billion).}
    \setlength{\tabcolsep}{0.14cm}
    \begin{tabular}{l|ll|llllccll}
    \hline
        Model & PS & T & WSC &WIC & RTE & CB & MULTIRC  & BOOLQ & COPA  & AVG \\ \hline
        GPT-Neo & 0.13 & 300 & 36.54  & 50.00  & 54.87  & 41.07  & 0.84  & 61.71  & 64.00  & 44.15  \\ 
        OPT & 0.16 & 300 & 36.54  & 50.00  & 49.82  & 21.43  & 1.36  & 55.47  & 66.00  & 40.09  \\ 
        Pythia & 0.16 & 300 & 36.54  & 50.16  & 52.71  & 41.07  & 2.52  & 55.08  & 65.00  &  43.30  \\ 
        HGRN & 0.15 & 100 & 38.46  & 51.10  & 56.68  & 42.86  & 1.47  & 59.91  & 65.00  &  45.07  \\ \hline
        OPT & 0.35 & 300 & 36.54  & 50.00  & 51.99  & 46.43  & 1.36  & 57.74  & 72.00  & 45.15  \\ 
        Pythia & 0.4 & 300 & 57.69  & 50.31  & 52.71  & 35.71  & 1.68  & 60.40  & 70.00  & 46.93 \\ 
        BLOOM & 0.56 & 350 & 40.38  & 50.00  & 52.71  & 41.07  & 1.05  & 55.14  & 61.00  & 43.05  \\ 
        HGRN & 0.35 & 100 & 38.46  & 50.16  & 52.71  & 51.79  & 1.99  & 59.05  & 73.00  & 46.74 \\ \hline
        GPT-Neo & 1.3 & 300 & 36.54  & 50.00  & 60.29  & 44.64  & 1.99  & 61.99  & 69.00  & 46.35  \\ 
        OPT & 1.3 & 300 & 37.50  & 51.10  & 51.99  & 41.07  & 3.15  & 57.77  & 79.00  & 45.94  \\ 
        Pythia & 1.4 & 300 & 36.54  & 50.00  & 53.07  & 35.71  & 0.94  & 60.73  & 72.00  & 44.14 \\ 
        BLOOM & 1.1 & 350 & 36.54  & 50.00  & 52.71  & 41.07  & 0.73  & 59.08  & 68.00  &  44.02 \\ 
        HGRN & 1 & 100 & 40.38  & 50.78  & 53.43  & 42.86  & 3.04  & 58.69  & 70.00  & 45.60 \\ \hline
    \end{tabular}
\end{table}

\paragraph{Long Range Arena}
LRA \cite{tay2020long} is proposed as a comprehensive evaluation for assessing the performances of models in processing long-term dependencies in various sequential modeling tasks.
We show a performance comparison between {\net} and existing methods in Table~\ref{table: lra res}.
{\net} achieves comparable results with other SOTA methods.

\paragraph{Image Classification}
The image classification results on the ImageNet-1K dataset are presented in Table~\ref{table imagenet1k}. Notably, with comparable parameter sizes, our proposed {\net} model demonstrates superior performance compared to previous methods such as TNN and the vanilla transformer.
It demonstrates the capability of \net~in modeling visual modalities.

\begin{table}[t]
\caption{\textbf{Performances Comparison on the Long Range Arena benchmark.} The proposed {\net} achieves the best performances and outperforms all competing methods.}
    \centering
    \setlength{\tabcolsep}{0.17cm}
    \begin{tabular}{l|cccccc|l}
    \hline
        Model & ListOps & Text & Retrieval & Image & Pathfinder & Path-X & AVG. \\ \hline
        Transformer~\cite{vaswani2017attention} & 38.37 & 61.95 & 80.69 & 40.57 & 65.26 & - & 47.81  \\ 
        cosFormer~\cite{zhen2022cosformer} & 36.50 & 67.70 & 83.15 & 51.23 & 71.96 & - & 51.76  \\ 
        FLASH~\cite{hua2022transformer} & 38.70 & 64.10 & 86.10 & 47.40 & 70.25 & - & 51.09  \\ 
        S4~\cite{gu2022efficiently} & 59.60 & 86.82 & 90.90 & 88.65 & 94.20 & 96.35 & 86.09  \\ 
        DSS\_softmax~\cite{dss} & 60.60 & 84.80 & 87.80 & 85.70 & 84.60 & 87.80 & 81.88  \\ 
        DSSEXP~\cite{dss} & 59.70 & 84.60 & 87.60 & 84.90 & 84.70 & 85.60 & 81.18  \\ 
        DSSEXP-NO-SCALE~\cite{dss} & 59.30 & 82.40 & 86.00 & 81.20 & 81.30 & - & 65.03  \\ 
        TNN~\cite{qin2023toeplitz} & 61.04 & 87.90 & 90.97 & 88.24 & 93.00 & 96.10 & 86.21  \\ 

        S5~\cite{s5} & 62.15 & 89.31 & 91.4 & 88 & 95.33 & 98.56 & 87.46 \\
        Mega~\cite{mega} & 63.14 & 90.43 & 91.25 & 90.44 & 96.01 & 97.98 & 88.21 \\
        SGConv~\cite{Li2022WhatMC} & 61.45 & 89.2 & 91.11 & 87.97 & 95.46 & 97.83 & 87.17 \\ 
        LRU~\cite{lru} & 60.20 & 89.40 & 89.90 & 89.00 & 95.10 & 94.20 & 86.30  \\ \hline
        {\net} & 59.95 & 88.14 & 94.23 & 88.69 & 92.92 & 97.50 & 86.91 \\ \hline
    \end{tabular}
    \label{table: lra res}
\end{table}

\begin{table}[h]
\centering
\small
\vspace{-3mm}
\caption{\textbf{Performances comparison of image classification on ImageNet-1k.}
{\net} performs favorably than competing methods with similar parameter sizes.}
\label{table imagenet1k}
 \setlength{\tabcolsep}{0.7cm}{
\begin{tabular}{c|cccc}
\hline
                    & \multicolumn{2}{c|}{DeiT-Tiny}                        & \multicolumn{2}{c}{DeiT-Small}   \\ \hline
Model               & \multicolumn{1}{c|}{Top1 Acc} & \multicolumn{1}{c|}{Param (M)} & \multicolumn{1}{c|}{Top1 Acc} & Parma (M) \\ \hline
Deit             & 72.20                    & 5.7                        & 79.90                    & 22.0  \\ \cline{1-1}
TNN                 & 72.29                    & 6.4                        & 79.20                    & 23.4  \\ \cline{1-1}
{\net} & 74.40                    & 6.1                        & 80.09                    & 23.7  \\ \hline
\end{tabular}}
\end{table}
\subsection{Ablation Study}
\begin{figure}[t]
     \centering
     \caption{\textbf{Visualization of forget rates.} We plot the forget rates of layers 5 and 6 on a model trained on language modeling tasks.}
     \vspace{-3mm}
     \begin{subfigure}[b]{0.3\textwidth}
         \centering
         \caption{With lower bound}
         \includegraphics[width=\textwidth]{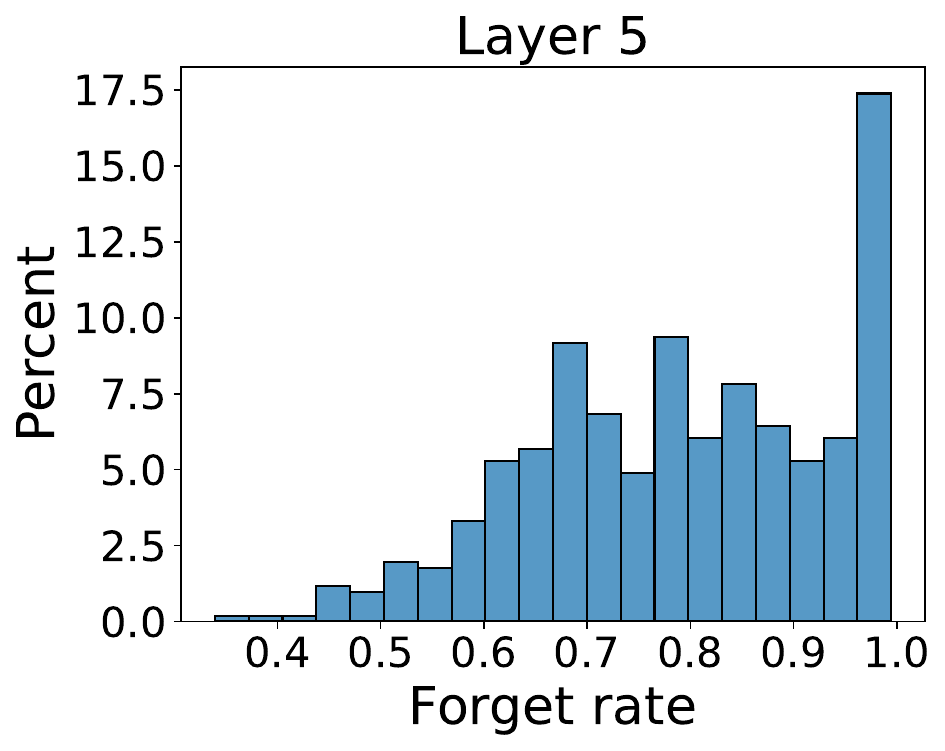}       
     \end{subfigure}
     \hfill
     \begin{subfigure}[b]{0.3\textwidth}
         \centering
         \caption{Without lower bound}
         \includegraphics[width=\textwidth]
         {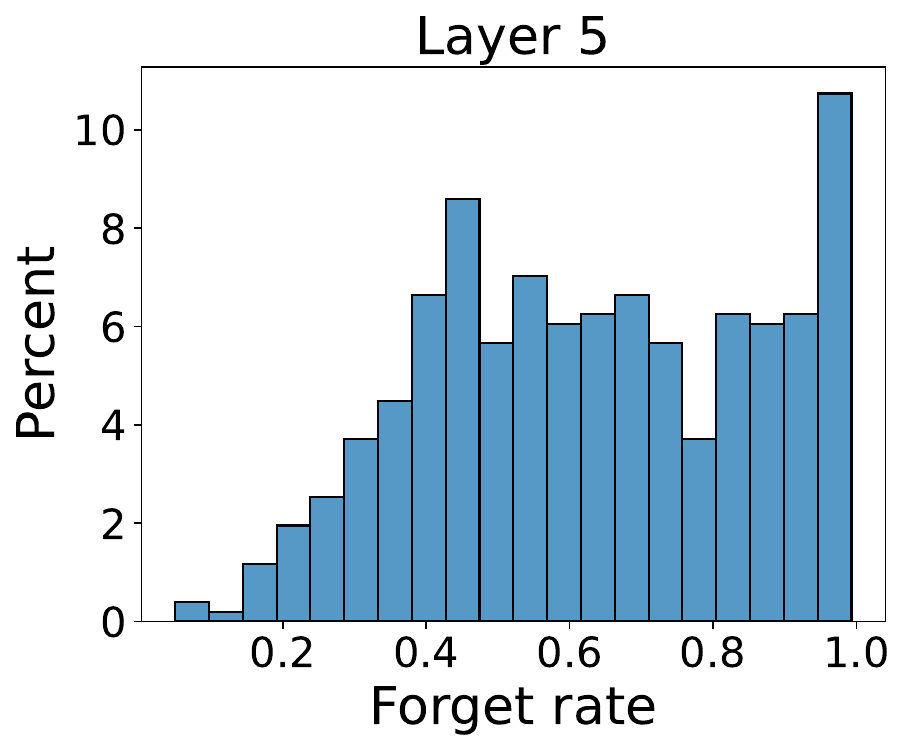}
     \end{subfigure} 
      \hfill
     \begin{subfigure}[b]{0.3\textwidth}
         \centering
         \caption{LRU}
         \includegraphics[width=\textwidth]
         {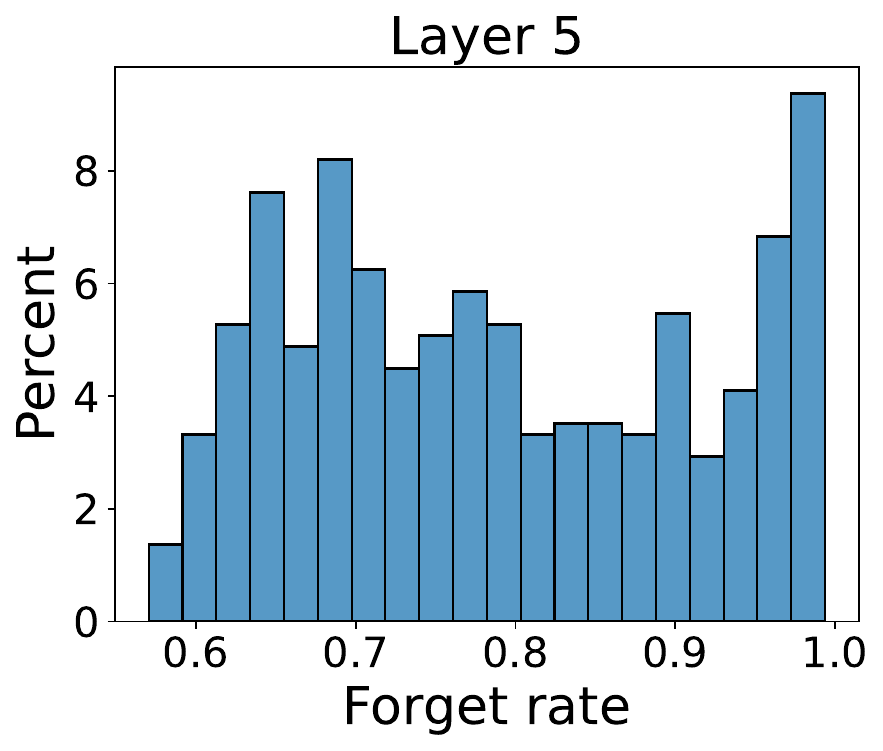}
     \end{subfigure}
     \begin{subfigure}[b]{0.3\textwidth}
         \centering
         \includegraphics[width=\textwidth]{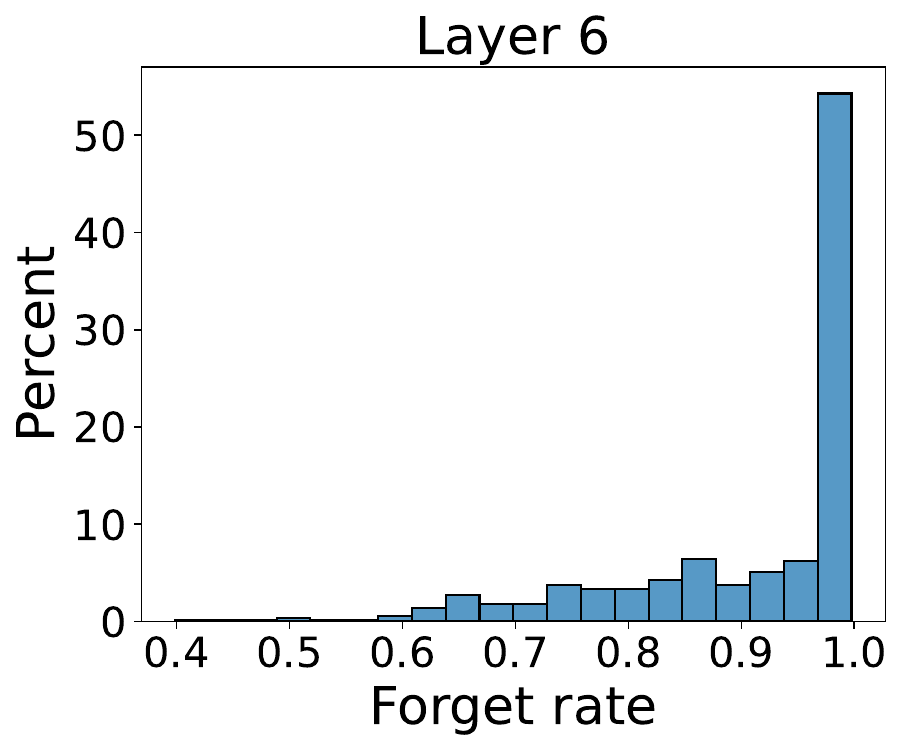}         
     \end{subfigure}
     \hfill
     \begin{subfigure}[b]{0.3\textwidth}
         \centering
         \includegraphics[width=\textwidth]
         {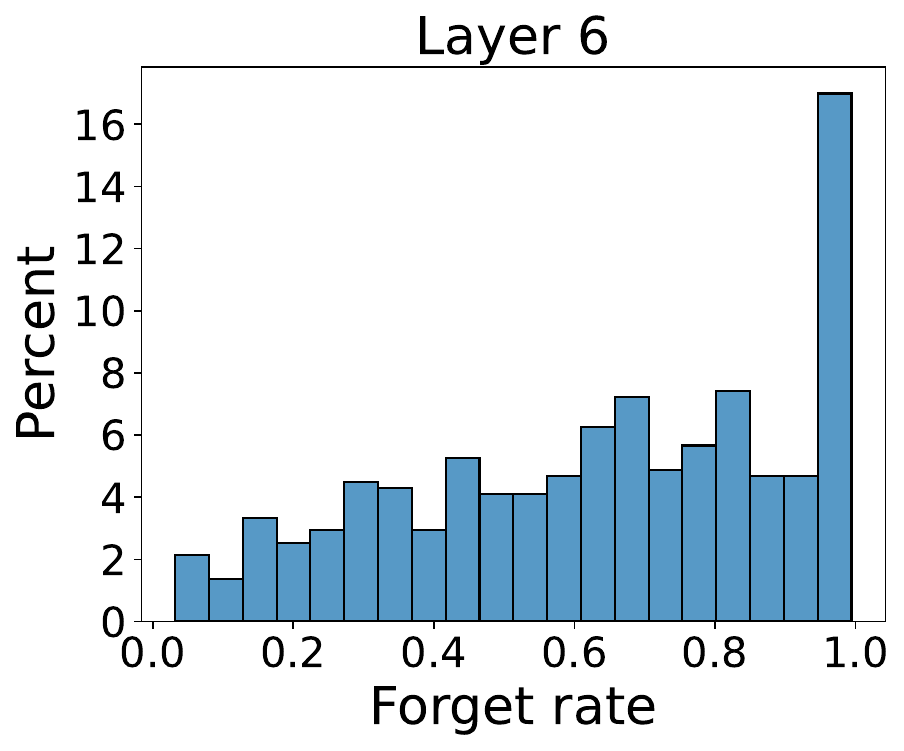}
     \end{subfigure} 
      \hfill
     \begin{subfigure}[b]{0.3\textwidth}
         \centering
         \includegraphics[width=\textwidth]
         {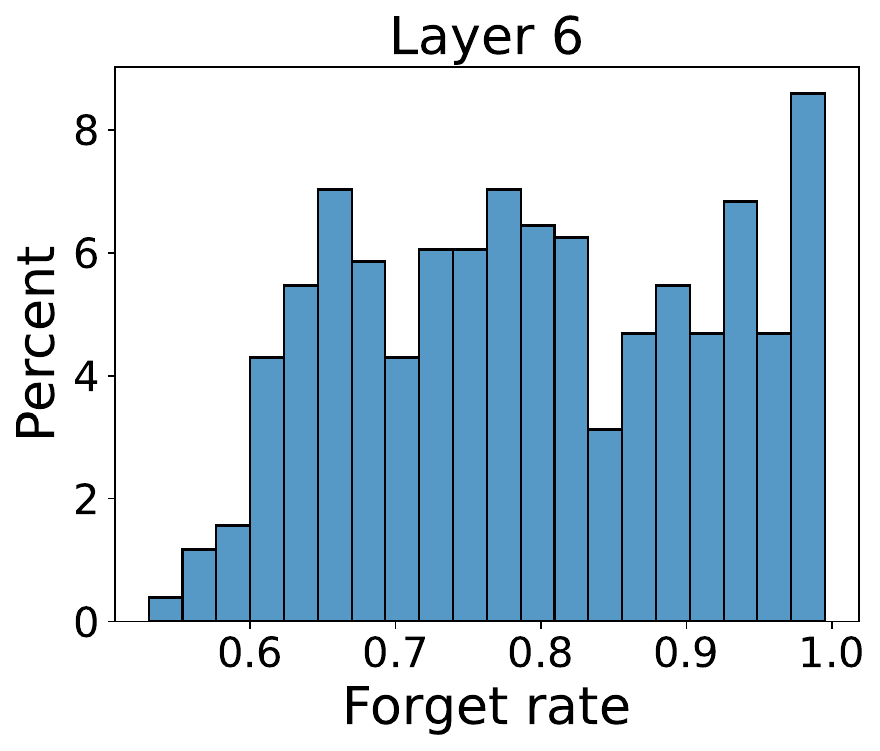}
     \end{subfigure}
     \newline
     \label{fig:hist-p1}
     \vspace{-6mm}
\end{figure}

\begin{wraptable}[7]{r}{0.5\textwidth}
\label{ablation:pile}
\small
\vspace{-14mm}
    \caption{\textbf{Forget gate ablation} on an autoregressive language model. The only lower bound means using a data-independent gate like LRU.}
    \centering
    \label{abl:forget}
    \setlength{\tabcolsep}{0.7cm}
    \begin{tabular}{l|ll}
    \hline
        Model & PPL$\downarrow$  \\ \hline
                LRU w forget gate & 4.92  \\ 
    LRU & 5.07  \\      
        {\net} only lower bound & 4.84  \\ 
        {\net} w/o forget gate & 57.42 \\ 
          \net & {4.14}  \\     
        \hline
        \end{tabular}
\end{wraptable}

We conducted ablation studies in the smallest-scaled setting (i.e., TNN\cite{qin2023toeplitz}'s setting on WikiText103 dataset) to thoroughly verify the effectiveness of each of our proposed components in \net. Experiments were conducted on the Pile dataset using a 1b model with 10b tokens for the forget gate experiment.

\paragraph{The influence of forget gate}
In table~\ref{abl:forget}, we demonstrate the role of forget gate. From table~\ref{abl:forget}, we observe that removing the forget gate significantly decreases the performance of \net, while adding a forget gate to LRU improves performance. On the other hand, using a data-independent forget gate (only lower bound) leads to lower performance compared to a data-dependent forget gate.

\paragraph{The influence of input gate and output gate}
\begin{wraptable}[8]{r}{0.5\textwidth}
\centering
\vspace{-4mm}
\caption{\textbf{Ablations of gates on autoregressive language modeling.} w/o input gate means to remove the $1 - \lambda_t$ term. w/o out\_gate means remove the left branch of {\name} in figure~\ref{fig: network}.}
            \label{table gate and norm}
            \setlength{\tabcolsep}{10mm}
            {          
          \begin{tabular}{l|l}
            \hline
                 Model & PPL$\downarrow$ \\ \hline
                w/o input gate  & 25.03 \\ 
                w/o output gate &  25.50 \\   
                 \net & 24.14 \\ \hline
            \end{tabular}
            \label{dpb:struct}
            }
\end{wraptable}
Table.~\ref{table gate and norm} validates the effectiveness of using output gates and tying input and forget gates. w/o input gate means to remove the $1-\lambda_t$ term. w/o output gate means remove the left branch of {\net} in figure~\ref{fig: network}. Our design achieves the best performance.

\paragraph{The influence of lower bounds in forget gate values}
\begin{wraptable}[13]{r}{0.5\textwidth}
\centering
\vspace{-4mm}
\caption{\textbf{Lower bound ablation} on autoregressive language modeling. A random lower bound means the lower bound in each layer is independent. Decrease lower bound means the lower bound is monotonically decreasing with respect to layer $k$, only the lower bound means the forget rate is independent of input.}
\vspace{-2mm}
            \label{table lower bound ablation}
            \setlength{\tabcolsep}{8mm}{
            \begin{tabular}{l|l}
            \hline
                Model & PPL $\downarrow$ \\ \hline
                w/o lower bound & 24.71 \\ 
                random lower bound & 24.60 \\ 
                decrease lower bound & 24.63 \\ 
                only lower bound & 27.70 \\ 
                \net & 24.14 \\ \hline
            \end{tabular} 
            }
\end{wraptable}
We demonstrate the effectiveness of introducing a lower bound in Table~\ref{table lower bound ablation} and Table~\ref{table lower bound on long sequence}. From Table~\ref{table lower bound ablation}, we observe that gating (i.e., without lower bound) is more critical than the lower bound (i.e., only lower bound). Combining gating and the lower bound consistently provides benefits, but the most significant improvement arises from the monotonically increasing lower bound. This aligns with the intuition that lower layers should primarily focus on nearby tokens, while upper layers should attend more broadly to capture long-term dependencies~\cite{poli2023hyena}.

Table~\ref{table lower bound on long sequence} highlights the essential role of the lower bound in long sequence processing tasks. In these tasks, the model's performance is notably poor and sometimes fails to converge without the lower bound. It is worth noting that language modeling tasks do not require extensive long-term dependencies, which explains why the model performs well even without the lower bound. However, in the task of LRA, the ability to capture long-term dependencies is crucial for achieving satisfactory performance. 

\paragraph{The influence of complex-valued recurrence}

\begin{wraptable}[11]{r}{0.5\textwidth}
\centering
\vspace{-4mm}
\caption{\textbf{Ablations of complex-valued recurrence on autoregressive language modeling.} w/o complex means remove theta, data-dependent theta means theta is dependent on the input, this makes the matrix $\Lambda$ not a Toeplitz matrix, which can not capture relative information.}
            \label{table complex}
            \setlength{\tabcolsep}{10mm}{
           \begin{tabular}{l|l}
            \hline
                 Model & PPL$\downarrow$ \\ \hline
                w/o complex & 25.34 \\ 
                data dependent $\theta$ & 28.74 \\ 
                 \net & 24.14 \\ \hline
            \end{tabular}
            }
\end{wraptable}
Table~\ref{table complex} validates the utility of incorporating complex values in element-wise linear recurrence. 
Additionally, the experiments show that the phase argument $\theta$ should not be data-dependent.

\subsection{Analysis on forget gate values}
We present the distributions of forget gate values across layers for different methods in Table~\ref{table layer decay rate} and visualize the histogram of each layer in Figure~\ref{fig:hist-p1}, trained on the autoregressive language modeling task. The results demonstrate that the addition of lower bounds effectively increases the average forget gate values in higher layers (5-6). Notably, the medium forget gate values in the highest layer reach 0.98, enabling the modeling of long-term dependencies.

It is interesting to note that the average forget gate values of the LRU model consistently exceed those of our variant model without lower bounds, as per their eigenvalues. However, despite this, the language modeling performance of LRU is lower than that of our variant. Specifically, LRU scored 24.71, while our variant scored 31.12. This suggests that using data-dependent gates to selectively retain relevant information is advantageous, rather than relying on data-independent forget gate values across all time steps.

\begin{table}[!ht]
\vspace{-3mm}
\caption{\textbf{Forget gate values of different methods on language modeling tasks.} In each layer, we counted the mean and median of forget gate values.}
    \centering
    \setlength{\tabcolsep}{0.32cm}
    \begin{tabular}{l|cc|cc|cc}
    \hline
              & ours & ours & w/o lower bound & w/o lower bound & LRU & LRU \\ \hline
        Layer & mean & median & mean & median & mean & median \\ \hline
        1 & 0.48  & 0.47  & 0.52  & 0.50  & 0.75  & 0.72  \\ 
        2 & 0.55  & 0.52  & 0.59  & 0.55  & 0.78  & 0.75  \\ 
        3 & 0.60  & 0.57  & 0.58  & 0.56  & 0.78  & 0.76  \\ 
        4 & 0.68  & 0.64  & 0.58  & 0.55  & 0.79  & 0.78  \\ 
        5 & 0.79  & 0.80  & 0.63  & 0.63  & 0.79  & 0.77  \\ 
        6 & 0.91  & 0.98  & 0.63  & 0.67  & 0.79  & 0.79 \\ \hline
    \end{tabular}
    \label{table layer decay rate}
\end{table}

\begin{table}[!ht]
\vspace{-5mm}
\caption{\textbf{Lower bound ablation on LRA.} We verify the importance of lower bounds in long-sequence modeling capabilities.}
    \centering
    \setlength{\tabcolsep}{0.25cm}
    \begin{tabular}{l|llllll|l}
    \hline
         Model & ListOps &  Text & Retrieval & Image & Pathfinder & Path-X & AVG \\ \hline

        w/o lower bound & 51.41  & 87.79  & 88.71  & 80.17  & -  & -  & 51.53\\ 
             \net& 59.95  & 88.14  & 94.23  & 88.69  & 92.92  & 97.50  & 86.91  \\ \hline
    \end{tabular}
    \label{table lower bound on long sequence}
\end{table}

\section{Conclusion}
In this work, we have shown that gated linear RNNs could obtain impressive performance across different tasks and modalities without compromising efficiency. We highlighted the significance of the forget gate for linear RNNs in language modeling and emphasized the importance of an additive lower bound on forget gate values for modeling long-term dependencies.  

\section*{Acknowledgement}
This work is partially supported by the National Key R\&D Program of China (NO.2022ZD0160100).

\section*{Limitations and broader impact}
Our empirical evaluation of {\net} remains on a smaller scale compared to other large-scale models. Potentially negative social consequences include the misuse of brain models for unsuitable purposes or applications, which must be prohibited by appropriate rules. In the era of large language models, the inference cost is the key limitation of transformer-based models. RNNs provide a solution with lower inference costs. This could potentially lead to a significant evolution in the field.

\newpage

\bibliography{neurips_2023}

\begin{thebibliography}{10}\itemsep=-1pt

\bibitem{akbari2021vatt}
Hassan Akbari, Liangzhe Yuan, Rui Qian, Wei-Hong Chuang, Shih-Fu Chang, Yin Cui, and Boqing Gong.
\newblock Vatt: Transformers for multimodal self-supervised learning from raw video, audio and text.
\newblock {\em arXiv preprint arXiv:2104.11178}, 2021.

\bibitem{DBLP:conf/icml/ArjovskySB16}
Mart{\'{\i}}n Arjovsky, Amar Shah, and Yoshua Bengio.
\newblock Unitary evolution recurrent neural networks.
\newblock In Maria{-}Florina Balcan and Kilian~Q. Weinberger, editors, {\em Proceedings of the 33nd International Conference on Machine Learning, {ICML} 2016, New York City, NY, USA, June 19-24, 2016}, volume~48 of {\em {JMLR} Workshop and Conference Proceedings}, pages 1120--1128. JMLR.org, 2016.

\bibitem{vivit}
Anurag Arnab, Mostafa Dehghani, Georg Heigold, Chen Sun, Mario Lu{\v{c}}i{\'c}, and Cordelia Schmid.
\newblock Vivit: A video vision transformer.
\newblock In {\em Proceedings of the IEEE/CVF International Conference on Computer Vision}, pages 6836--6846, 2021.

\bibitem{strongtypedrnn}
David Balduzzi and Muhammad Ghifary.
\newblock Strongly-typed recurrent neural networks.
\newblock In Maria{-}Florina Balcan and Kilian~Q. Weinberger, editors, {\em Proceedings of the 33nd International Conference on Machine Learning, {ICML} 2016, New York City, NY, USA, June 19-24, 2016}, volume~48 of {\em {JMLR} Workshop and Conference Proceedings}, pages 1292--1300. JMLR.org, 2016.

\bibitem{dilatedrnn}
Shiyu Chang, Yang Zhang, Wei Han, Mo Yu, Xiaoxiao Guo, Wei Tan, Xiaodong Cui, Michael Witbrock, Mark~A. Hasegawa{-}Johnson, and Thomas~S. Huang.
\newblock Dilated recurrent neural networks.
\newblock In Isabelle Guyon, Ulrike von Luxburg, Samy Bengio, Hanna~M. Wallach, Rob Fergus, S.~V.~N. Vishwanathan, and Roman Garnett, editors, {\em Advances in Neural Information Processing Systems 30: Annual Conference on Neural Information Processing Systems 2017, December 4-9, 2017, Long Beach, CA, {USA}}, pages 77--87, 2017.

\bibitem{cho-etal-2014-learning}
Kyunghyun Cho, Bart van Merri{\"e}nboer, Caglar Gulcehre, Dzmitry Bahdanau, Fethi Bougares, Holger Schwenk, and Yoshua Bengio.
\newblock Learning phrase representations using {RNN} encoder{--}decoder for statistical machine translation.
\newblock In {\em Proceedings of the 2014 Conference on Empirical Methods in Natural Language Processing ({EMNLP})}, pages 1724--1734, Doha, Qatar, Oct. 2014. Association for Computational Linguistics.

\bibitem{choromanski2020rethinking}
Krzysztof Choromanski, Valerii Likhosherstov, David Dohan, Xingyou Song, Andreea Gane, Tamas Sarlos, Peter Hawkins, Jared Davis, Afroz Mohiuddin, Lukasz Kaiser, et~al.
\newblock Rethinking attention with performers.
\newblock {\em arXiv preprint arXiv:2009.14794}, 2020.

\bibitem{DBLP:conf/iclr/ChungAB17}
Junyoung Chung, Sungjin Ahn, and Yoshua Bengio.
\newblock Hierarchical multiscale recurrent neural networks.
\newblock In {\em 5th International Conference on Learning Representations, {ICLR} 2017, Toulon, France, April 24-26, 2017, Conference Track Proceedings}. OpenReview.net, 2017.

\bibitem{gru}
Junyoung Chung, {\c{C}}aglar G{\"{u}}l{\c{c}}ehre, KyungHyun Cho, and Yoshua Bengio.
\newblock Empirical evaluation of gated recurrent neural networks on sequence modeling.
\newblock {\em CoRR}, abs/1412.3555, 2014.

\bibitem{dao2022flashattention}
Tri Dao, Daniel~Y. Fu, Stefano Ermon, Atri Rudra, and Christopher R{\'e}.
\newblock Flash{A}ttention: Fast and memory-efficient exact attention with {IO}-awareness.
\newblock In {\em Advances in Neural Information Processing Systems}, 2022.

\bibitem{h3}
Tri Dao, Daniel~Y. Fu, Khaled~Kamal Saab, Armin~W. Thomas, Atri Rudra, and Christopher R{\'{e}}.
\newblock Hungry hungry hippos: Towards language modeling with state space models.
\newblock {\em CoRR}, abs/2212.14052, 2022.

\bibitem{devlin-etal-2019-bert}
Jacob Devlin, Ming-Wei Chang, Kenton Lee, and Kristina Toutanova.
\newblock {BERT}: Pre-training of deep bidirectional transformers for language understanding.
\newblock In {\em Proceedings of the 2019 Conference of the North {A}merican Chapter of the Association for Computational Linguistics: Human Language Technologies, Volume 1 (Long and Short Papers)}, pages 4171--4186, Minneapolis, Minnesota, June 2019. Association for Computational Linguistics.

\bibitem{vit}
Alexey Dosovitskiy, Lucas Beyer, Alexander Kolesnikov, Dirk Weissenborn, Xiaohua Zhai, Thomas Unterthiner, Mostafa Dehghani, Matthias Minderer, Georg Heigold, Sylvain Gelly, et~al.
\newblock An image is worth 16x16 words: Transformers for image recognition at scale.
\newblock {\em arXiv preprint arXiv:2010.11929}, 2020.

\bibitem{simplelongconv}
Daniel~Y. Fu, Elliot~L. Epstein, Eric Nguyen, Armin~W. Thomas, Michael Zhang, Tri Dao, Atri Rudra, and Christopher R{\'{e}}.
\newblock Simple hardware-efficient long convolutions for sequence modeling.
\newblock {\em CoRR}, abs/2302.06646, 2023.

\bibitem{pile}
Leo Gao, Stella Biderman, Sid Black, Laurence Golding, Travis Hoppe, Charles Foster, Jason Phang, Horace He, Anish Thite, Noa Nabeshima, Shawn Presser, and Connor Leahy.
\newblock The {P}ile: An 800gb dataset of diverse text for language modeling.
\newblock {\em arXiv preprint arXiv:2101.00027}, 2020.

\bibitem{eval-harness}
Leo Gao, Jonathan Tow, Stella Biderman, Sid Black, Anthony DiPofi, Charles Foster, Laurence Golding, Jeffrey Hsu, Kyle McDonell, Niklas Muennighoff, Jason Phang, Laria Reynolds, Eric Tang, Anish Thite, Ben Wang, Kevin Wang, and Andy Zou.
\newblock A framework for few-shot language model evaluation.
\newblock Zenodo, Sept. 2021.

\bibitem{DBLP:journals/neco/GersSC00}
Felix~A. Gers, J{\"{u}}rgen Schmidhuber, and Fred~A. Cummins.
\newblock Learning to forget: Continual prediction with {LSTM}.
\newblock {\em Neural Comput.}, 12(10):2451--2471, 2000.

\bibitem{gong21b_interspeech}
Yuan Gong, Yu-An Chung, and James Glass.
\newblock {AST: Audio Spectrogram Transformer}.
\newblock In {\em Proc. Interspeech 2021}, pages 571--575, 2021.

\bibitem{Greff2015LSTMAS}
Klaus Greff, Rupesh~Kumar Srivastava, Jan Koutn{\'i}k, Bas~R. Steunebrink, and J{\"u}rgen Schmidhuber.
\newblock Lstm: A search space odyssey.
\newblock {\em IEEE Transactions on Neural Networks and Learning Systems}, 28:2222--2232, 2015.

\bibitem{s4d}
Albert Gu, Karan Goel, Ankit Gupta, and Christopher R{\'{e}}.
\newblock On the parameterization and initialization of diagonal state space models.
\newblock In {\em NeurIPS}, 2022.

\bibitem{s4}
Albert Gu, Karan Goel, and Christopher R{\'{e}}.
\newblock Efficiently modeling long sequences with structured state spaces.
\newblock In {\em The Tenth International Conference on Learning Representations, {ICLR} 2022, Virtual Event, April 25-29, 2022}. OpenReview.net, 2022.

\bibitem{gu2022efficiently}
Albert Gu, Karan Goel, and Christopher R\'e.
\newblock Efficiently modeling long sequences with structured state spaces.
\newblock In {\em The International Conference on Learning Representations ({ICLR})}, 2022.

\bibitem{gu-improving}
Albert Gu, {\c{C}}aglar G{\"{u}}l{\c{c}}ehre, Thomas Paine, Matt Hoffman, and Razvan Pascanu.
\newblock Improving the gating mechanism of recurrent neural networks.
\newblock In {\em Proceedings of the 37th International Conference on Machine Learning, {ICML} 2020, 13-18 July 2020, Virtual Event}, volume 119 of {\em Proceedings of Machine Learning Research}, pages 3800--3809. {PMLR}, 2020.

\bibitem{DBLP:conf/nips/GuJGSDRR21}
Albert Gu, Isys Johnson, Karan Goel, Khaled Saab, Tri Dao, Atri Rudra, and Christopher R{\'{e}}.
\newblock Combining recurrent, convolutional, and continuous-time models with linear state space layers.
\newblock In Marc'Aurelio Ranzato, Alina Beygelzimer, Yann~N. Dauphin, Percy Liang, and Jennifer~Wortman Vaughan, editors, {\em Advances in Neural Information Processing Systems 34: Annual Conference on Neural Information Processing Systems 2021, NeurIPS 2021, December 6-14, 2021, virtual}, pages 572--585, 2021.

\bibitem{2110.13985}
Albert Gu, Isys Johnson, Karan Goel, Khaled Saab, Tri Dao, Atri Rudra, and Christopher Ré.
\newblock Combining recurrent, convolutional, and continuous-time models with linear state-space layers, 2021.

\bibitem{dss}
Ankit Gupta, Albert Gu, and Jonathan Berant.
\newblock Diagonal state spaces are as effective as structured state spaces, 2022.

\bibitem{DBLP:journals/corr/abs-2212-00768}
Ankit Gupta, Harsh Mehta, and Jonathan Berant.
\newblock Simplifying and understanding state space models with diagonal linear rnns.
\newblock {\em CoRR}, abs/2212.00768, 2022.

\bibitem{liquids4}
Ramin Hasani, Mathias Lechner, Tsun-Hsuan Wang, Makram Chahine, Alexander Amini, and Daniela Rus.
\newblock Liquid structural state-space models.
\newblock In {\em The Eleventh International Conference on Learning Representations}, 2023.

\bibitem{unified}
Hongyu He and Marko Kabic.
\newblock A unified view of long-sequence models towards modeling million-scale dependencies.
\newblock {\em CoRR}, abs/2302.06218, 2023.

\bibitem{Hochreiter2001GradientFI}
Sepp Hochreiter and Yoshua Bengio.
\newblock Gradient flow in recurrent nets: the difficulty of learning long-term dependencies.
\newblock 2001.

\bibitem{hua2022transformer}
Weizhe Hua, Zihang Dai, Hanxiao Liu, and Quoc~V Le.
\newblock Transformer quality in linear time.
\newblock {\em arXiv preprint arXiv:2202.10447}, 2022.

\bibitem{huang2023encoding}
Feiqing Huang, Kexin Lu, Yuxi CAI, Zhen Qin, Yanwen Fang, Guangjian Tian, and Guodong Li.
\newblock Encoding recurrence into transformers.
\newblock In {\em The Eleventh International Conference on Learning Representations}, 2023.

\bibitem{Hunter1986TheEW}
J.~Stuart Hunter.
\newblock The exponentially weighted moving average.
\newblock {\em Journal of Quality Technology}, 18:203--210, 1986.

\bibitem{xfmrsarernns}
Angelos Katharopoulos, Apoorv Vyas, Nikolaos Pappas, and Fran{\c{c}}ois Fleuret.
\newblock Transformers are rnns: Fast autoregressive transformers with linear attention.
\newblock In {\em Proceedings of the 37th International Conference on Machine Learning, {ICML} 2020, 13-18 July 2020, Virtual Event}, volume 119 of {\em Proceedings of Machine Learning Research}, pages 5156--5165. {PMLR}, 2020.

\bibitem{katharopoulos2020transformers}
Angelos Katharopoulos, Apoorv Vyas, Nikolaos Pappas, and Fran{\c{c}}ois Fleuret.
\newblock Transformers are rnns: Fast autoregressive transformers with linear attention.
\newblock In {\em International Conference on Machine Learning}, pages 5156--5165. PMLR, 2020.

\bibitem{ke2021rethinking}
Guolin Ke, Di He, and Tie-Yan Liu.
\newblock Rethinking positional encoding in language pre-training.
\newblock In {\em International Conference on Learning Representations}, 2021.

\bibitem{DBLP:conf/icml/KoutnikGGS14}
Jan Koutn{\'{\i}}k, Klaus Greff, Faustino~J. Gomez, and J{\"{u}}rgen Schmidhuber.
\newblock A clockwork {RNN}.
\newblock In {\em Proceedings of the 31th International Conference on Machine Learning, {ICML} 2014, Beijing, China, 21-26 June 2014}, volume~32 of {\em {JMLR} Workshop and Conference Proceedings}, pages 1863--1871. JMLR.org, 2014.

\bibitem{Le2015ASW}
Quoc~V. Le, Navdeep Jaitly, and Geoffrey~E. Hinton.
\newblock A simple way to initialize recurrent networks of rectified linear units.
\newblock {\em ArXiv}, abs/1504.00941, 2015.

\bibitem{lee-thorp-etal-2022-fnet}
James Lee-Thorp, Joshua Ainslie, Ilya Eckstein, and Santiago Ontanon.
\newblock {FN}et: Mixing tokens with {F}ourier transforms.
\newblock In {\em Proceedings of the 2022 Conference of the North American Chapter of the Association for Computational Linguistics: Human Language Technologies}, pages 4296--4313, Seattle, United States, July 2022. Association for Computational Linguistics.

\bibitem{lei-etal-2018-simple}
Tao Lei, Yu Zhang, Sida~I. Wang, Hui Dai, and Yoav Artzi.
\newblock Simple recurrent units for highly parallelizable recurrence.
\newblock In {\em Proceedings of the 2018 Conference on Empirical Methods in Natural Language Processing}, pages 4470--4481, Brussels, Belgium, Oct.-Nov. 2018. Association for Computational Linguistics.

\bibitem{Li2022WhatMC}
Yuhong Li, Tianle Cai, Yi Zhang, De huai Chen, and Debadeepta Dey.
\newblock What makes convolutional models great on long sequence modeling?
\newblock {\em ArXiv}, abs/2210.09298, 2022.

\bibitem{liu2021pay}
Hanxiao Liu, Zihang Dai, David So, and Quoc~V Le.
\newblock Pay attention to mlps.
\newblock {\em Advances in Neural Information Processing Systems}, 34:9204--9215, 2021.

\bibitem{liu2019roberta}
Yinhan Liu, Myle Ott, Naman Goyal, Jingfei Du, Mandar Joshi, Danqi Chen, Omer Levy, Mike Lewis, Luke Zettlemoyer, and Veselin Stoyanov.
\newblock Roberta: A robustly optimized bert pretraining approach.
\newblock {\em arXiv preprint arXiv:1907.11692}, 2019.

\bibitem{liu2022neural}
Zexiang Liu, Dong Li, Kaiyue Lu, Zhen Qin, Weixuan Sun, Jiacheng Xu, and Yiran Zhong.
\newblock Neural architecture search on efficient transformers and beyond.
\newblock In {\em arXiv preprint arXiv:2207.13955}, 2022.

\bibitem{lu2022linear}
Kaiyue Lu, Zexiang Liu, Jianyuan Wang, Weixuan Sun, Zhen Qin, Dong Li, Xuyang Shen, Hui Deng, Xiaodong Han, Yuchao Dai, et~al.
\newblock Linear video transformer with feature fixation.
\newblock {\em arXiv preprint arXiv:2210.08164}, 2022.

\bibitem{mega}
Xuezhe Ma, Chunting Zhou, Xiang Kong, Junxian He, Liangke Gui, Graham Neubig, Jonathan May, and Luke Zettlemoyer.
\newblock Mega: Moving average equipped gated attention.
\newblock {\em CoRR}, abs/2209.10655, 2022.

\bibitem{mao-2022-fine}
Huanru~Henry Mao.
\newblock Fine-tuning pre-trained transformers into decaying fast weights.
\newblock In {\em Proceedings of the 2022 Conference on Empirical Methods in Natural Language Processing}, pages 10236--10242, Abu Dhabi, United Arab Emirates, Dec. 2022. Association for Computational Linguistics.

\bibitem{iclr18}
Eric Martin and Chris Cundy.
\newblock Parallelizing linear recurrent neural nets over sequence length.
\newblock In {\em 6th International Conference on Learning Representations, {ICLR} 2018, Vancouver, BC, Canada, April 30 - May 3, 2018, Conference Track Proceedings}. OpenReview.net, 2018.

\bibitem{gss}
Harsh Mehta, Ankit Gupta, Ashok Cutkosky, and Behnam Neyshabur.
\newblock Long range language modeling via gated state spaces.
\newblock {\em CoRR}, abs/2206.13947, 2022.

\bibitem{merity2017pointer}
Stephen Merity, Caiming Xiong, James Bradbury, and Richard Socher.
\newblock Pointer sentinel mixture models.
\newblock {\em 5th International Conference on Learning Representations, {ICLR}, Toulon, France}, 2017.

\bibitem{miao2015eesen}
Yajie Miao, Mohammad Gowayyed, and Florian Metze.
\newblock Eesen: End-to-end speech recognition using deep rnn models and wfst-based decoding.
\newblock In {\em 2015 IEEE Workshop on Automatic Speech Recognition and Understanding (ASRU)}, pages 167--174. IEEE, 2015.

\bibitem{DBLP:journals/corr/abs-2307-11888}
Antonio Orvieto, Soham De, {\c{C}}aglar G{\"{u}}l{\c{c}}ehre, Razvan Pascanu, and Samuel~L. Smith.
\newblock On the universality of linear recurrences followed by nonlinear projections.
\newblock {\em CoRR}, abs/2307.11888, 2023.

\bibitem{lru}
Antonio Orvieto, Samuel~L. Smith, Albert Gu, Anushan Fernando, {\c{C}}aglar G{\"{u}}l{\c{c}}ehre, Razvan Pascanu, and Soham De.
\newblock Resurrecting recurrent neural networks for long sequences.
\newblock {\em CoRR}, abs/2303.06349, 2023.

\bibitem{paszke2019pytorch}
Adam Paszke, Sam Gross, Francisco Massa, Adam Lerer, James Bradbury, Gregory Chanan, Trevor Killeen, Zeming Lin, Natalia Gimelshein, Luca Antiga, et~al.
\newblock Pytorch: An imperative style, high-performance deep learning library.
\newblock {\em Advances in neural information processing systems}, 32, 2019.

\bibitem{rwkv}
Bo Peng, Eric Alcaide, Quentin Anthony, Alon Albalak, Samuel Arcadinho, Huanqi Cao, Xin Cheng, Michael Chung, Matteo Grella, Kranthi Kiran~G. V., Xuzheng He, Haowen Hou, Przemyslaw Kazienko, Jan Kocon, Jiaming Kong, Bartlomiej Koptyra, Hayden Lau, Krishna Sri~Ipsit Mantri, Ferdinand Mom, Atsushi Saito, Xiangru Tang, Bolun Wang, Johan~S. Wind, Stanislaw Wozniak, Ruichong Zhang, Zhenyuan Zhang, Qihang Zhao, Peng Zhou, Jian Zhu, and Rui{-}Jie Zhu.
\newblock {RWKV:} reinventing rnns for the transformer era.
\newblock {\em CoRR}, abs/2305.13048, 2023.

\bibitem{rfa}
Hao Peng, Nikolaos Pappas, Dani Yogatama, Roy Schwartz, Noah~A. Smith, and Lingpeng Kong.
\newblock Random feature attention.
\newblock In {\em 9th International Conference on Learning Representations, {ICLR} 2021, Virtual Event, Austria, May 3-7, 2021}. OpenReview.net, 2021.

\bibitem{hyena}
Michael Poli, Stefano Massaroli, Eric Nguyen, Daniel~Y. Fu, Tri Dao, Stephen Baccus, Yoshua Bengio, Stefano Ermon, and Christopher R{\'{e}}.
\newblock Hyena hierarchy: Towards larger convolutional language models.
\newblock {\em CoRR}, abs/2302.10866, 2023.

\bibitem{poli2023hyena}
Michael Poli, Stefano Massaroli, Eric Nguyen, Daniel~Y Fu, Tri Dao, Stephen Baccus, Yoshua Bengio, Stefano Ermon, and Christopher R{\'e}.
\newblock Hyena hierarchy: Towards larger convolutional language models.
\newblock {\em arXiv preprint arXiv:2302.10866}, 2023.

\bibitem{qin2023toeplitz}
Zhen Qin, Xiaodong Han, Weixuan Sun, Bowen He, Dong Li, Dongxu Li, Yuchao Dai, Lingpeng Kong, and Yiran Zhong.
\newblock Toeplitz neural network for sequence modeling.
\newblock In {\em The Eleventh International Conference on Learning Representations (ICLR)}, 2023.

\bibitem{qin-etal-2022-devil}
Zhen Qin, Xiaodong Han, Weixuan Sun, Dongxu Li, Lingpeng Kong, Nick Barnes, and Yiran Zhong.
\newblock The devil in linear transformer.
\newblock In {\em Proceedings of the 2022 Conference on Empirical Methods in Natural Language Processing}, pages 7025--7041, Abu Dhabi, United Arab Emirates, Dec. 2022. Association for Computational Linguistics.

\bibitem{qin2023scaling}
Zhen Qin, Dong Li, Weigao Sun, Weixuan Sun, Xuyang Shen, Xiaodong Han, Yunshen Wei, Baohong Lv, Fei Yuan, Xiao Luo, Yu Qiao, and Yiran Zhong.
\newblock Scaling transnormer to 175 billion parameters.
\newblock {\em arXiv}, 2023.

\bibitem{zhen2022cosformer}
Zhen Qin, Weixuan Sun, Hui Deng, Dongxu Li, Yunshen Wei, Baohong Lv, Junjie Yan, Lingpeng Kong, and Yiran Zhong.
\newblock cosformer: Rethinking softmax in attention.
\newblock In {\em ICLR}, 2022.

\bibitem{qin2023linearized}
Zhen Qin, Weixuan Sun, Kaiyue Lu, Hui Deng, Dongxu Li, Xiaodong Han, Yuchao Dai, Lingpeng Kong, and Yiran Zhong.
\newblock Linearized relative positional encoding.
\newblock {\em Transactions on Machine Learning Research}, 2023.

\bibitem{qin-etal-2023-linear}
Zhen Qin and Yiran Zhong.
\newblock Accelerating toeplitz neural network with constant-time inference complexity.
\newblock In {\em Proceedings of the 2023 Conference on Empirical Methods in Natural Language Processing}. Association for Computational Linguistics, Dec. 2023.

\bibitem{salman2015weather}
Afan~Galih Salman, Bayu Kanigoro, and Yaya Heryadi.
\newblock Weather forecasting using deep learning techniques.
\newblock In {\em 2015 international conference on advanced computer science and information systems (ICACSIS)}, pages 281--285. Ieee, 2015.

\bibitem{Schlag2021LinearTA}
Imanol Schlag, Kazuki Irie, and J{\"u}rgen Schmidhuber.
\newblock Linear transformers are secretly fast weight programmers.
\newblock In {\em International Conference on Machine Learning}, 2021.

\bibitem{Schmidhuber1992LearningTC}
J{\"u}rgen Schmidhuber.
\newblock Learning to control fast-weight memories: An alternative to dynamic recurrent networks.
\newblock {\em Neural Computation}, 4:131--139, 1992.

\bibitem{selvin2017stock}
Sreelekshmy Selvin, R Vinayakumar, EA Gopalakrishnan, Vijay~Krishna Menon, and KP Soman.
\newblock Stock price prediction using lstm, rnn and cnn-sliding window model.
\newblock In {\em 2017 international conference on advances in computing, communications and informatics (icacci)}, pages 1643--1647. IEEE, 2017.

\bibitem{glu}
Noam Shazeer.
\newblock {GLU} variants improve transformer.
\newblock {\em CoRR}, abs/2002.05202, 2020.

\bibitem{onlstm}
Yikang Shen, Shawn Tan, Alessandro Sordoni, and Aaron~C. Courville.
\newblock Ordered neurons: Integrating tree structures into recurrent neural networks.
\newblock In {\em 7th International Conference on Learning Representations, {ICLR} 2019, New Orleans, LA, USA, May 6-9, 2019}. OpenReview.net, 2019.

\bibitem{s5}
Jimmy T.~H. Smith, Andrew Warrington, and Scott~W. Linderman.
\newblock Simplified state space layers for sequence modeling.
\newblock {\em CoRR}, abs/2208.04933, 2022.

\bibitem{rope}
Jianlin Su, Yu Lu, Shengfeng Pan, Bo Wen, and Yunfeng Liu.
\newblock Roformer: Enhanced transformer with rotary position embedding.
\newblock {\em CoRR}, abs/2104.09864, 2021.

\bibitem{sun2022locality}
Jingyu Sun, Guiping Zhong, Dinghao Zhou, Baoxiang Li, and Yiran Zhong.
\newblock Locality matters: A locality-biased linear attention for automatic speech recognition.
\newblock {\em arXiv preprint arXiv:2203.15609}, 2022.

\bibitem{Sun2023Tpami}
Weixuan Sun, Zhen Qin, Hui Deng, Jianyuan Wang, Yi Zhang, Kaihao Zhang, Nick Barnes, Stan Birchfield, Lingpeng Kong, and Yiran Zhong.
\newblock Vicinity vision transformer.
\newblock {\em IEEE Transactions on Pattern Analysis and Machine Intelligence}, 45(10):12635--12649, 2023.

\bibitem{DBLP:conf/iclr/TallecO18a}
Corentin Tallec and Yann Ollivier.
\newblock Can recurrent neural networks warp time?
\newblock In {\em 6th International Conference on Learning Representations, {ICLR} 2018, Vancouver, BC, Canada, April 30 - May 3, 2018, Conference Track Proceedings}. OpenReview.net, 2018.

\bibitem{tay2021synthesizer}
Yi Tay, Dara Bahri, Donald Metzler, Da-Cheng Juan, Zhe Zhao, and Che Zheng.
\newblock Synthesizer: Rethinking self-attention for transformer models.
\newblock In {\em International conference on machine learning}, pages 10183--10192. PMLR, 2021.

\bibitem{tay2020long}
Yi Tay, Mostafa Dehghani, Samira Abnar, Yikang Shen, Dara Bahri, Philip Pham, Jinfeng Rao, Liu Yang, Sebastian Ruder, and Donald Metzler.
\newblock Long range arena: A benchmark for efficient transformers.
\newblock In {\em International Conference on Learning Representations}, 2020.

\bibitem{lra}
Yi Tay, Mostafa Dehghani, Samira Abnar, Yikang Shen, Dara Bahri, Philip Pham, Jinfeng Rao, Liu Yang, Sebastian Ruder, and Donald Metzler.
\newblock Long range arena : {A} benchmark for efficient transformers.
\newblock In {\em 9th International Conference on Learning Representations, {ICLR} 2021, Virtual Event, Austria, May 3-7, 2021}. OpenReview.net, 2021.

\bibitem{pmlr-v139-touvron21a}
Hugo Touvron, Matthieu Cord, Matthijs Douze, Francisco Massa, Alexandre Sablayrolles, and Herve Jegou.
\newblock Training data-efficient image transformers \&amp; distillation through attention.
\newblock In {\em International Conference on Machine Learning}, volume 139, pages 10347--10357, July 2021.

\bibitem{forgetgate}
Jos van~der Westhuizen and Joan Lasenby.
\newblock The unreasonable effectiveness of the forget gate.
\newblock {\em CoRR}, abs/1804.04849, 2018.

\bibitem{vaswani2017attention}
Ashish Vaswani, Noam Shazeer, Niki Parmar, Jakob Uszkoreit, Llion Jones, Aidan~N Gomez, {\L}ukasz Kaiser, and Illia Polosukhin.
\newblock Attention is all you need.
\newblock {\em Advances in neural information processing systems}, 30, 2017.

\bibitem{pretrainingwoattn}
Junxiong Wang, Jing~Nathan Yan, Albert Gu, and Alexander~M. Rush.
\newblock Pretraining without attention.
\newblock {\em CoRR}, abs/2212.10544, 2022.

\bibitem{metaformer}
Weihao Yu, Mi Luo, Pan Zhou, Chenyang Si, Yichen Zhou, Xinchao Wang, Jiashi Feng, and Shuicheng Yan.
\newblock Metaformer is actually what you need for vision.
\newblock In {\em {IEEE/CVF} Conference on Computer Vision and Pattern Recognition, {CVPR} 2022, New Orleans, LA, USA, June 18-24, 2022}, pages 10809--10819. {IEEE}, 2022.

\bibitem{zhou2021informer}
Haoyi Zhou, Shanghang Zhang, Jieqi Peng, Shuai Zhang, Jianxin Li, Hui Xiong, and Wancai Zhang.
\newblock Informer: Beyond efficient transformer for long sequence time-series forecasting.
\newblock In {\em Proceedings of the AAAI conference on artificial intelligence}, volume~35, pages 11106--11115, 2021.

\bibitem{zhu2021longshort}
Chen Zhu, Wei Ping, Chaowei Xiao, Mohammad Shoeybi, Tom Goldstein, Anima Anandkumar, and Bryan Catanzaro.
\newblock Long-short transformer: Efficient transformers for language and vision.
\newblock In A. Beygelzimer, Y. Dauphin, P. Liang, and J.~Wortman Vaughan, editors, {\em Advances in Neural Information Processing Systems}, 2021.

\end{thebibliography}
\bibliographystyle{neurips_2023}

\clearpage

\section{Appendix}
In this appendix, we examine the extrapolation ability of \net~and provide the training and inference speed comparison of \net~and existing efficient sequence modeling methods. We also illustrate the forget rates of each layer on a trained language model of \net.
We also report the extrapolation ability of {{\net}} compared to previous methods in Table~\ref{extrapola}.

\subsection{Extrapolation test}
In this section, we tested {\net} 's extrapolation ability by directly inferring the model with a variety of sequence lengths. As shown in Table~\ref{extrapola}, our method has the ability to train short and test long.

\begin{sidewaystable}[t]
    \centering
    \caption{The extrapolation performance of competing methods. The best result is highlighted in \textbf{bold} and the second in \underline{underline}.$\downarrow$ means \textit{lower is better}.}
    \begin{tabular}{|l|l|l|l|l|l|l|l|l|l|l|l|l|l|l|l|l|}
    \hline
     \makecell[c]{\\Seqlen} & \makecell[c]{Transformer\\ PPL$\downarrow$} & \makecell[c]{LS\\ PPL$\downarrow$} & \makecell[c]{FLASH\\ PPL$\downarrow$} & \makecell[c]{1+elu\\ PPL$\downarrow$} & \makecell[c]{Performer\\ PPL$\downarrow$} & \makecell[c]{cosFormer\\ PPL$\downarrow$}  & \makecell[c]{gMLP\\ PPL$\downarrow$} & \makecell[c]{S4\\ PPL$\downarrow$} & \makecell[c]{DSS \\ PPL$\downarrow$}& \makecell[c]{GSS\\ PPL$\downarrow$} & \makecell[c]{ALiBi\\ PPL$\downarrow$} & \makecell[c]{TNN\\ PPL$\downarrow$} 
& \makecell[c]{LRU\\ PPL$\downarrow$}& \makecell[c]{\name\\ PPL$\downarrow$}\\
\hline
        512 & 24.78 & 24.05 & 24.69 & 28.05 & 63.16 & 27.06  & 29.13 & 30.74 & 41.07 & 39.66 & 24.15 & 24.67 & 31.12 & 24.85\\ 
        768 & 41.36 & 23.49 & 16950.45 & 47.35 & 159.74 & 32.90  & 1.34E+9 & 30.41 & 40.50 & 39.76 & 23.38 & 24.25 &30.72 &24.4\\
        1024 & 62.35 & 23.21 & 174165.47 & 70.47 & 504.30 & 55.28  & 8.93E+12 & 30.24 & 40.22 & 39.91 & 22.98 & 24.05& 30.5 &24.16\\
        1280 & 82.52 & 23.07 & 346502.88 & 91.88 & 1020.28 & 102.88  & 1.58E+15 & 30.15 & 40.03 & 40.82 & 22.74 & 23.91 & 30.38 &24.03\\ 
        1536 & 100.17 & 22.97 & 647788.12 & 111.56 & 1568.83 & 175.26 & 4.96E+16 & 30.08 & 39.94 & 41.04 & 22.57 & 23.83 &30.3 &23.94\\ 
        1792 & 118.42 & 22.97 & 1719873.5 & 129.92 & 2138.50 & 267.65 & 5.67E+17 & 30.04 & 39.85 & 41.08 & 22.52 & 23.79 & 30.24 &23.88 \\
        2048 & 133.44 & 22.99 & 6.25E+6 & 147.09 & 2693.89 & 368.02 & 3.59E+18 & 30.00 & 39.79 & 41.53 & 22.43 & 23.73 &30.19 & 23.82 \\
        3072 & 188.95 & 23.25 & 4.17E+10 & 206.88 & 4945.82 & 820.77  & 2.19E+20 & 29.91 & 39.64 & 44.08 & 22.24 & 23.63 & 30.09 &23.71 \\
        4096 & 246.06 & 23.83 & 2.67E+13 & 267.87 & 7170.91 & 1335.51  & 1.61E+21 & 29.88 & 39.59 & 48.27 & 22.17 & 23.58 & 30.04 &23.66\\
        5120 & 270.93 & 24.56 & 1.26E+15 & 299.31 & 8443.15 & 1735.50 & 5.08E+21 & 29.85 & 39.54 & 53.32 & 22.11 & 23.54 & 30.01 &23.62 \\
        6144 & 311.65 & 25.45 & 1.58E+16 & 352.62 & 10234.07 & 2146.19  & 1.16E+22 & 29.83 & 39.51 & 57.73 & 22.08 & 23.53 & 29.99 &23.6\\
        7168 & 346.58 & 26.42 & 8.11E+16 & 389.02 & 11420.56 & 2494.79 & 1.98E+22 & 29.82 & 39.49 & 60.25 & 22.07 & 23.51 & 29.97 & 23.58\\
        8192 & 372.18 & 27.11 & 3.40E+17 & 411.50 & 12557.09 & 2902.24  & 2.78E+22 & 29.82 & 39.49 & 63.36 & 22.05 & 23.51 & 29.97 & 23.58\\
        9216 & 387.29 & 28.78 & 1.22E+18 & 453.27 & 14847.66 & 3028.72  & 3.93E+22 & 29.80 & 39.46 & 74.92 & 22.03 & 23.49 & 29.96 & 23.56\\
        10240 & 395.94 & 30.13 & 4.03E+18 & 457.06 & 13623.83 & 3247.83 & 4.93E+22 & 29.79 & 39.45 & 81.87 & 22.02 & 23.48 & 29.94 & 23.55\\
        11264 & 426.54 & 31.14 & 1.07E+19 & 504.19 & 14661.77 & 3341.91 & 5.70E+22 & 29.79 & 39.46 & 87.67 & 22.00 & 23.48 & 29.94 & 23.55\\
        12288 & 463.50 & 33.21 & 2.52E+19 & 555.38 & 17959.85 & 3644.81  & 7.18E+22 & 29.79 & 39.44 & 92.11 & 22.00 & 23.48 & 29.94 & 23.55\\
        13312 & 506.35 & 34.72 & 4.96E+19 & 584.01 & 20026.35 & 3851.70 & 8.04E+22 & 29.78 & 39.43 & 96.00 & 22.00 & 23.47 & 29.93 & 23.54\\
        14336 & 486.86 & 36.05 & 1.28E+20 & 589.83 & 20971.31 & 3951.26 & 9.41E+22 & 29.78 & 39.43 & 101.47 & 21.99 & 23.46 & 29.92 &23.53\\  \hline
        Avg & 261.36 & 26.71 & 1.16E+19 & 299.86 & 8684.79 & 1764.75  & 2.41E+22 & 29.97 & 39.75 & 60.26 & \textbf{22.40} & \underline{23.70} & 30.17 & 23.80 \\ \hline
    
    \end{tabular}
    \label{extrapola}
\end{sidewaystable}

\subsection{Speed comparison}
In this section, we benchmark the speed of our method on the LRA benchmark. Our method achieves state-of-the-art training and inference speed.

\begin{table}[h]
    \centering
  \caption{Speed comparison on LRA benchmark. The 1K,...,5K represent the input sequence length. We mark it with - if a method is out of memory. The higher the better for all metrics. }
    \begin{tabular}{l|lllll|lllll}
    \hline
            & \multicolumn{5}{c|}{Train Speed(steps per second)$\uparrow$} & \multicolumn{5}{c}{Inference Speed(steps per second)$\uparrow$} \\ \hline
        Method & 1K & 2K & 3K & 4K & 5K & 1K & 2K & 3K & 4K & 5K \\
        Transformer~\cite{vaswani2017attention} &13.58	 & 4.84 & - & - & - & 23.67 &	8.22 & - & - & - \\ 
        Performer~\cite{ke2021rethinking}  & 18.40  & 10.77 & 7.66 & 6.30 & 5.64 & 30.04 & 17.36 & 12.80 & 10.55 & 9.52 \\ 
        LS~\cite{zhu2021longshort} & 20.29 & 11.24 & 8.05 & 6.51 & 5.89 & 39.05 & 21.11 & 15.02 & 12.6 & 11.66 \\
        Fnet~\cite{lee-thorp-etal-2022-fnet} & 25.19 & 15.62 & 11.24 & 9.41 & 8.18 & 48.81 & 27.89 & 19.52 & 16.27 & 14.46 \\ 
       cosFormer~\cite{zhen2022cosformer} & 22.00 & 12.80 & 9.47 & 7.93 & 7.13 & 39.05 & 22.31 & 16.62 & 13.95 & 12.60 \\ 
        S4~\cite{s4} & 13.13 & 7.33 & 4.91 & 3.84 & 3.04 & 30.04 & 16.27 & 10.85 & 8.58 & 6.79 \\ 
        FLASH~\cite{hua2022transformer} & 17.36 & 9.03 & 6.54 & 5.19 & 4.68 & 30.04 & 15.94 & 11.32 & 9.19 & 8.40 \\ 
        TNN~\cite{qin2023toeplitz} & 17.55 & 9.89 & 6.79 & 5.68 & 4.54 & 33.96 & 17.75 & 12.40 & 10.28 & 8.22 \\ 
         \name & 22.31 & 13.58 & 9.52 & 7.40 & 7.44 & 43.39 & 25.19 & 16.62 & 14.20 & 13.95 \\ \hline
    \end{tabular}
\end{table}

\subsection{Visualization}
\label{histplot}
In this section, we visualize the forget rates of each layer on a model trained on language modeling tasks. 

\begin{figure}
\label{fig:hist-p2}
     \centering
     \caption{Visualization forget rates in each layer.}
     \begin{subfigure}[b]{0.3\textwidth}
         \centering
         \caption{With lower bound}
         \includegraphics[width=\textwidth]{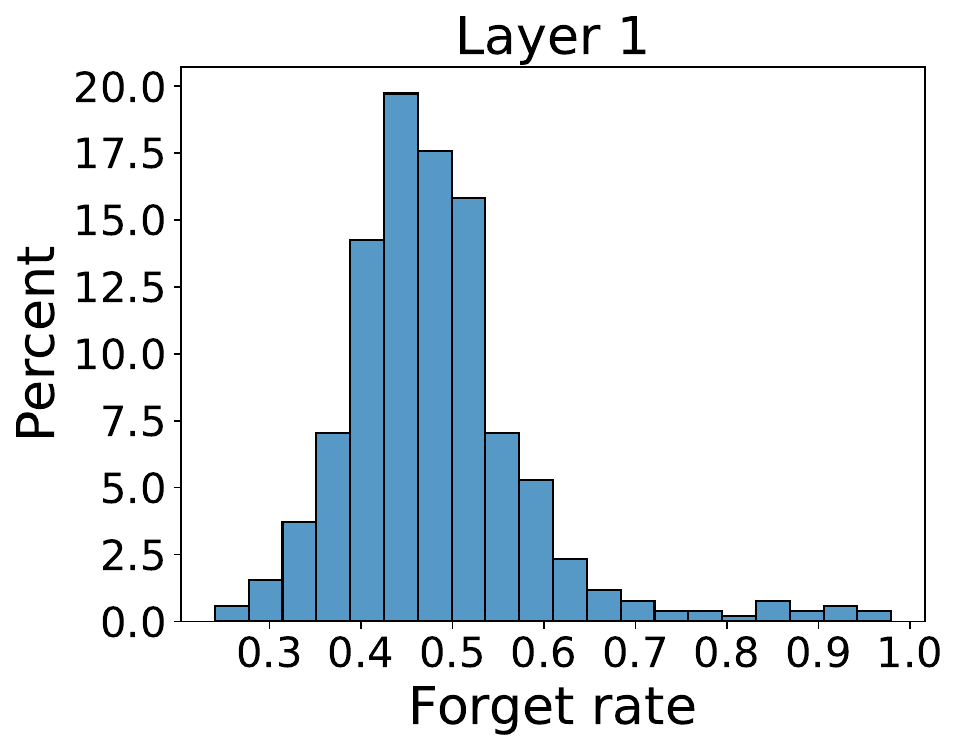}     
     \end{subfigure}
     \hfill
     \begin{subfigure}[b]{0.3\textwidth}
         \centering
         \caption{Without lower bound}
         \includegraphics[width=\textwidth]
         {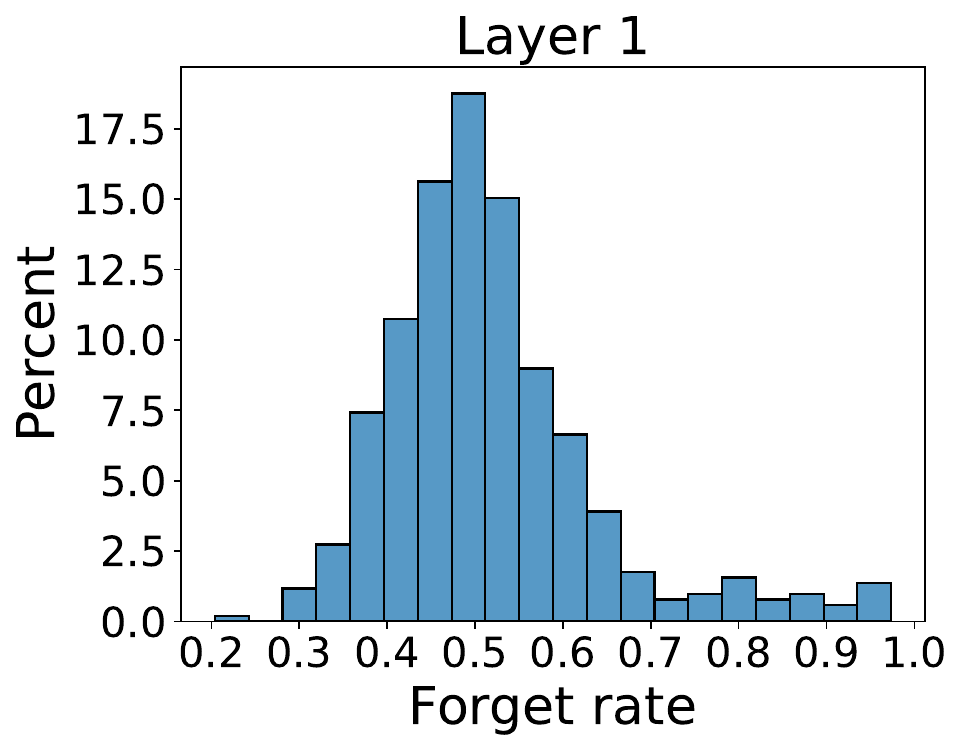}
     \end{subfigure} 
      \hfill
     \begin{subfigure}[b]{0.3\textwidth}
         \centering
         \caption{LRU}
         \includegraphics[width=\textwidth]
         {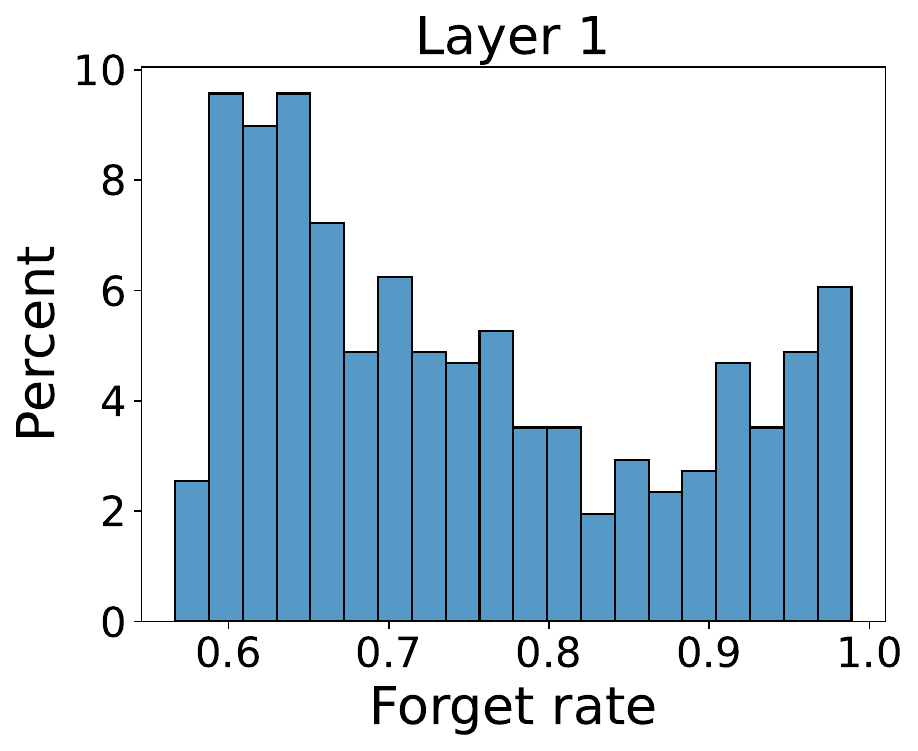}
     \end{subfigure}
     \newline
     \begin{subfigure}[b]{0.3\textwidth}
         \centering
         \includegraphics[width=\textwidth]{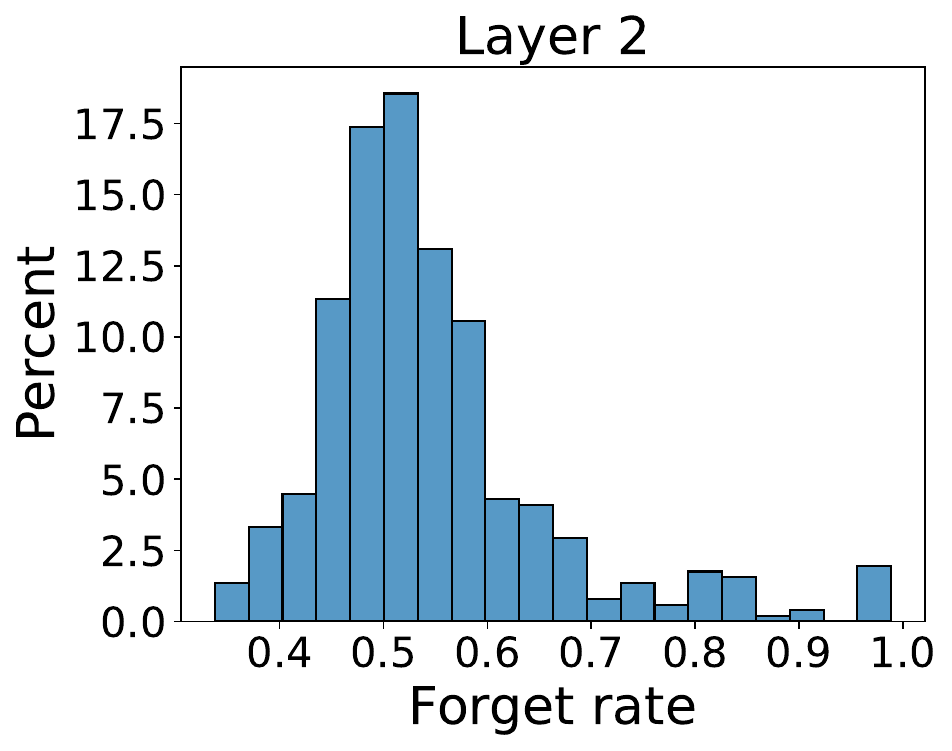}        
     \end{subfigure}
     \hfill
     \begin{subfigure}[b]{0.3\textwidth}
         \centering
         \includegraphics[width=\textwidth]
         {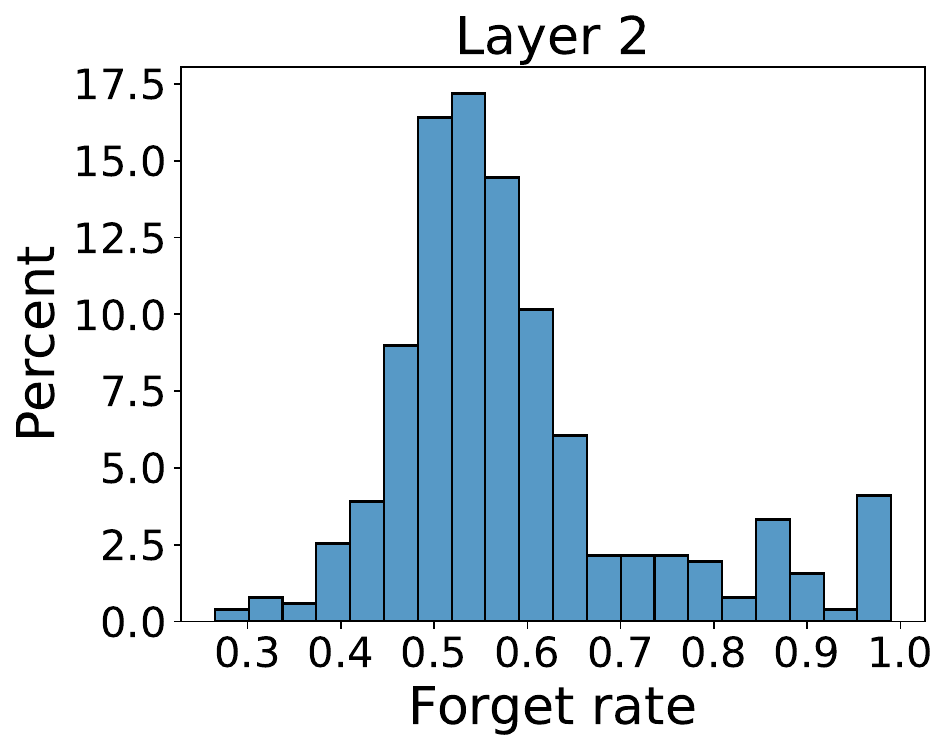}
     \end{subfigure} 
      \hfill
     \begin{subfigure}[b]{0.3\textwidth}
         \centering
         \includegraphics[width=\textwidth]
         {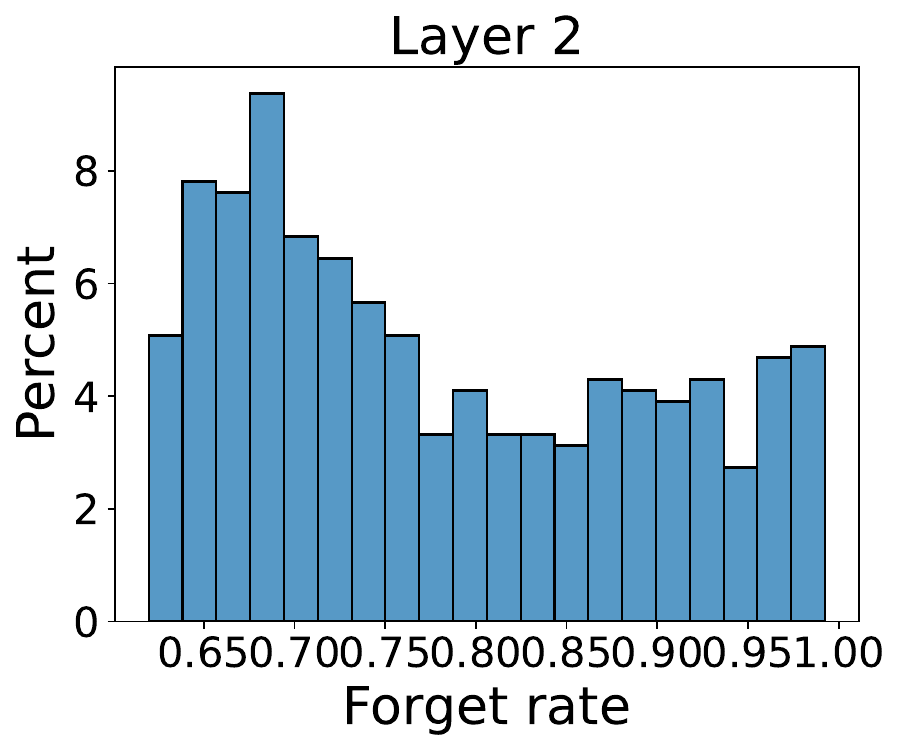}
     \end{subfigure}
     \newline
     \begin{subfigure}[b]{0.3\textwidth}
         \centering
         \includegraphics[width=\textwidth]{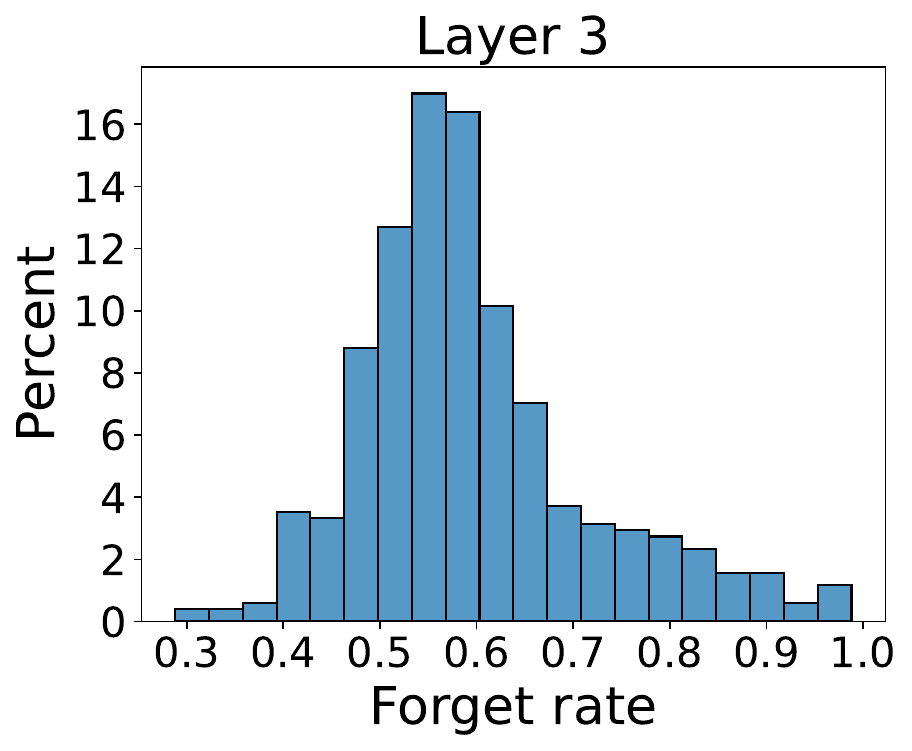}         
     \end{subfigure}
     \hfill
     \begin{subfigure}[b]{0.3\textwidth}
         \centering
         \includegraphics[width=\textwidth]
         {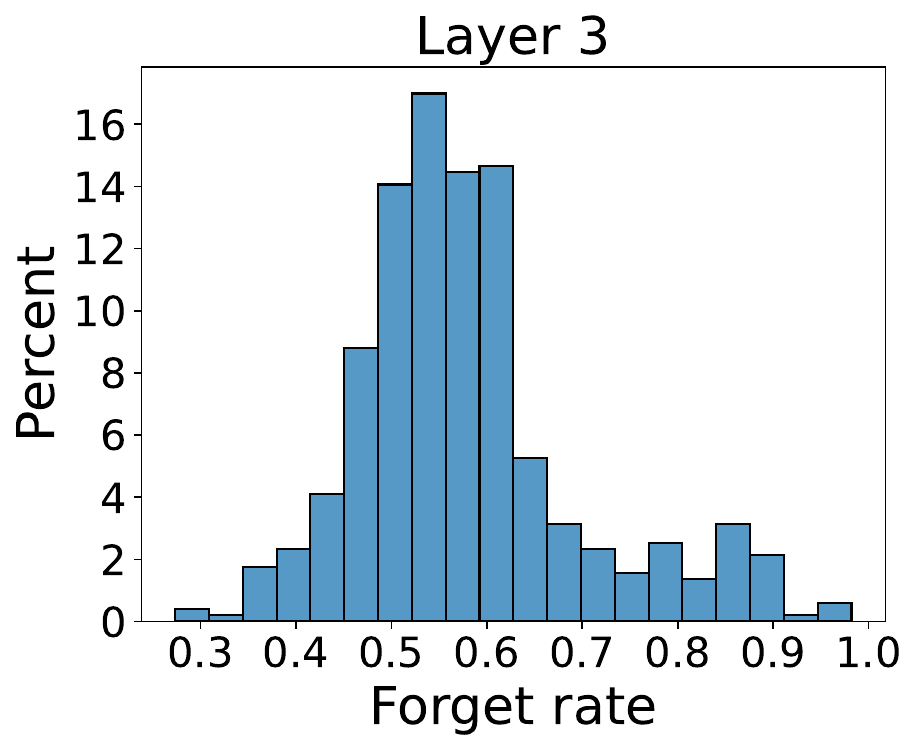}
     \end{subfigure} 
      \hfill
     \begin{subfigure}[b]{0.3\textwidth}
         \centering
         \includegraphics[width=\textwidth]
         {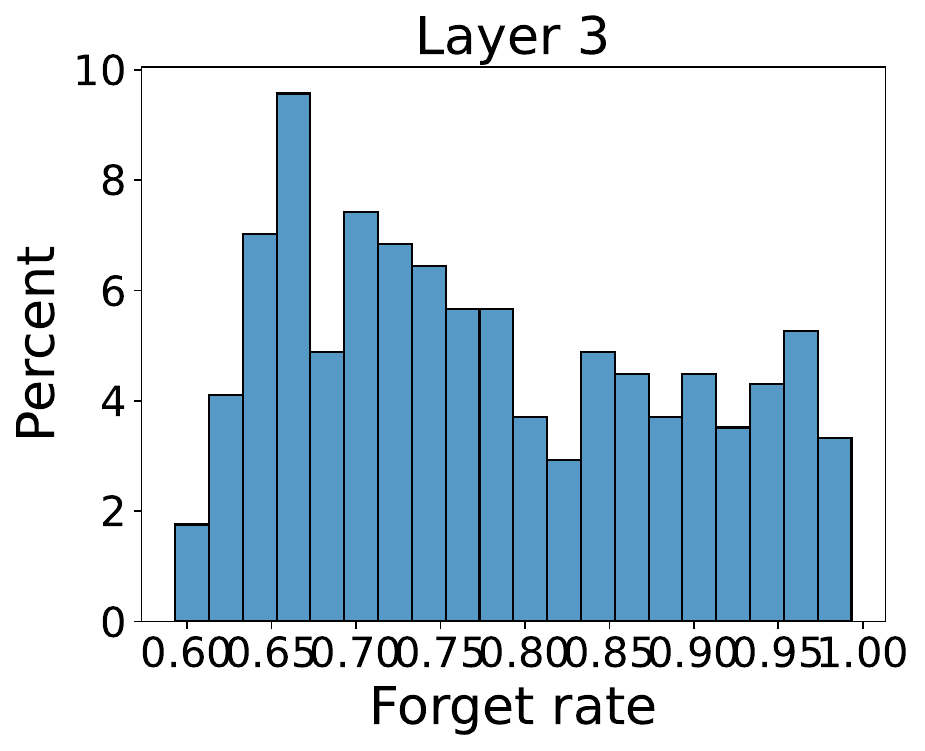}
     \end{subfigure}
     \newline
     \begin{subfigure}[b]{0.3\textwidth}
         \centering
         \includegraphics[width=\textwidth]{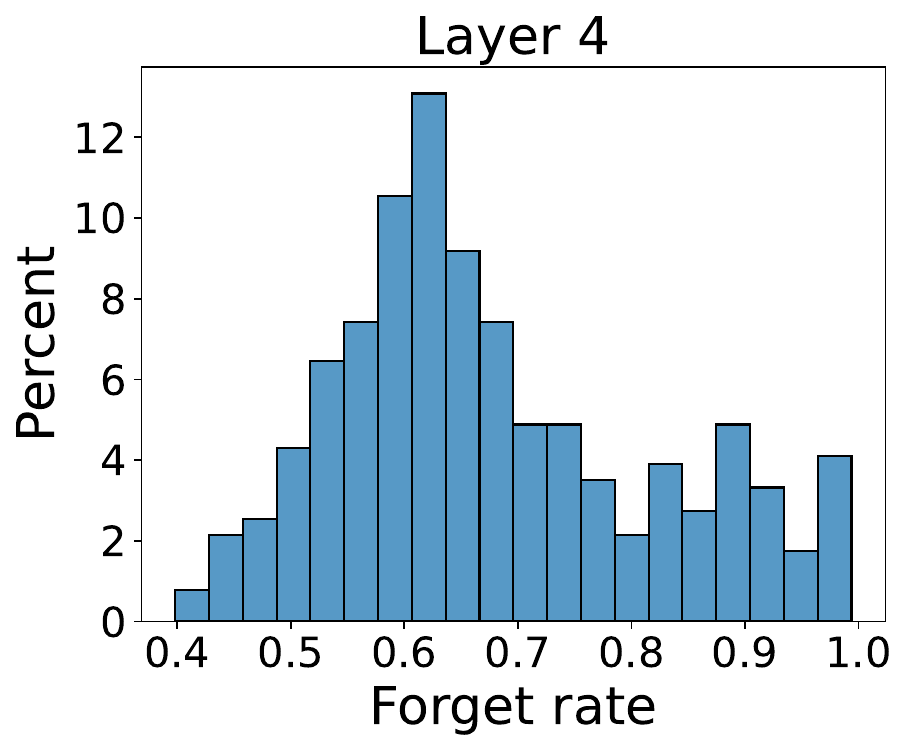}         
     \end{subfigure}
     \hfill
     \begin{subfigure}[b]{0.3\textwidth}
         \centering
         \includegraphics[width=\textwidth]
         {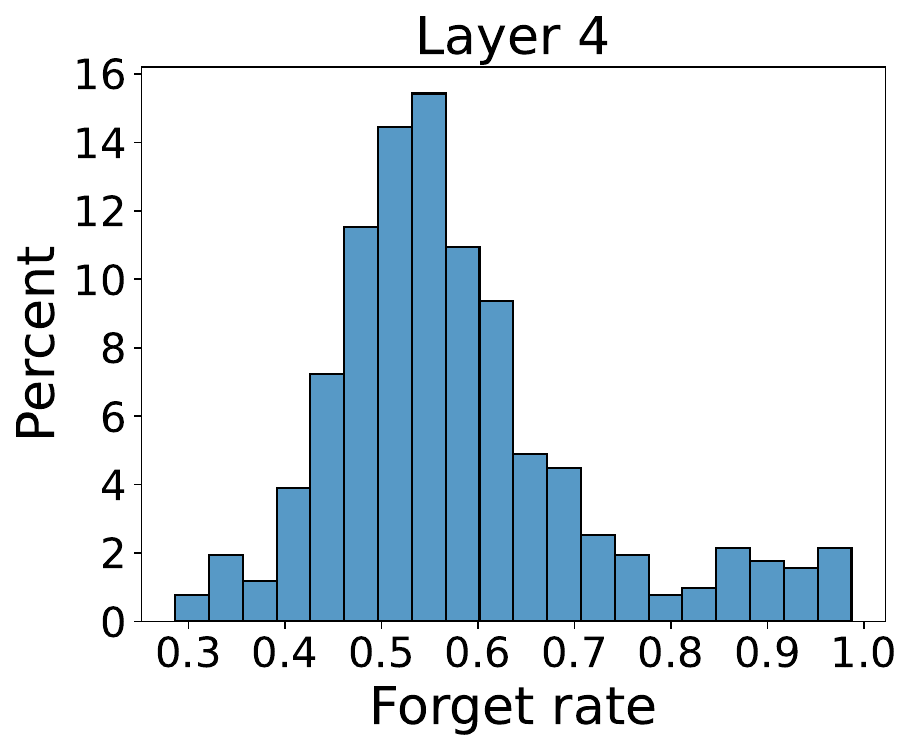}
     \end{subfigure} 
      \hfill
     \begin{subfigure}[b]{0.3\textwidth}
         \centering
         \includegraphics[width=\textwidth]
         {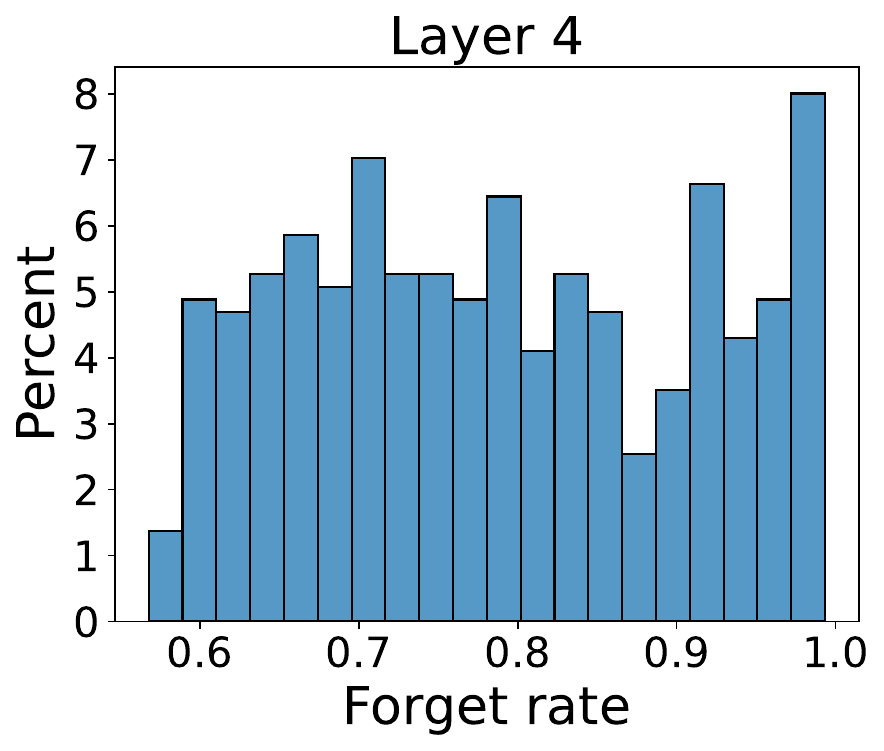}
     \end{subfigure}
     \newline
     \begin{subfigure}[b]{0.3\textwidth}
         \centering
         \includegraphics[width=\textwidth]{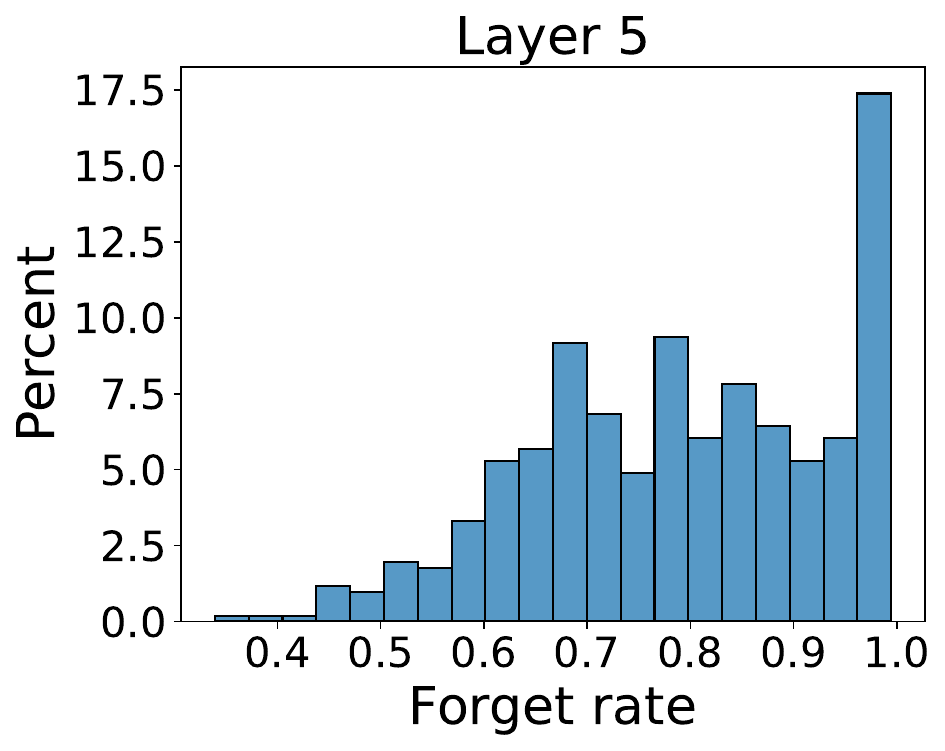}         
     \end{subfigure}
     \hfill
     \begin{subfigure}[b]{0.3\textwidth}
         \centering
         \includegraphics[width=\textwidth]
         {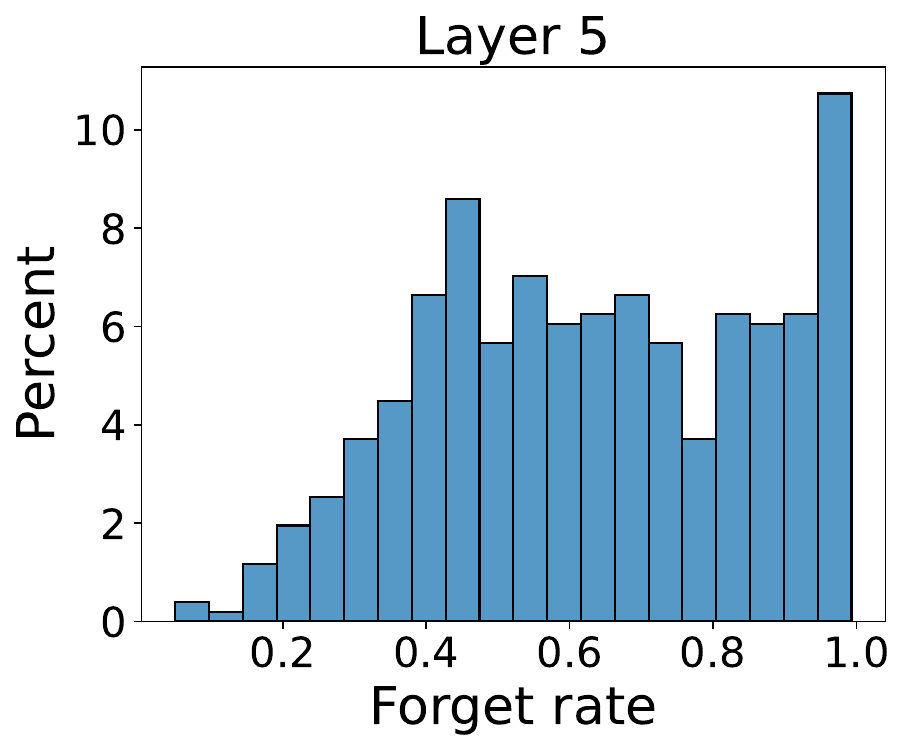}
     \end{subfigure} 
      \hfill
     \begin{subfigure}[b]{0.3\textwidth}
         \centering
         \includegraphics[width=\textwidth]
         {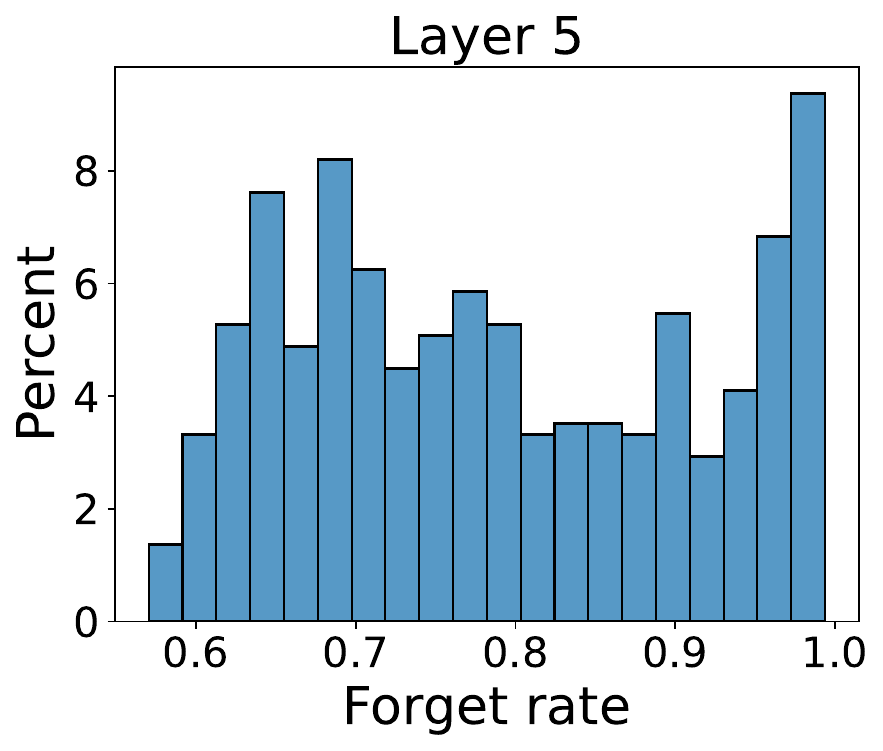}
     \end{subfigure}
     \newline
     \begin{subfigure}[b]{0.3\textwidth}
         \centering
         \includegraphics[width=\textwidth]{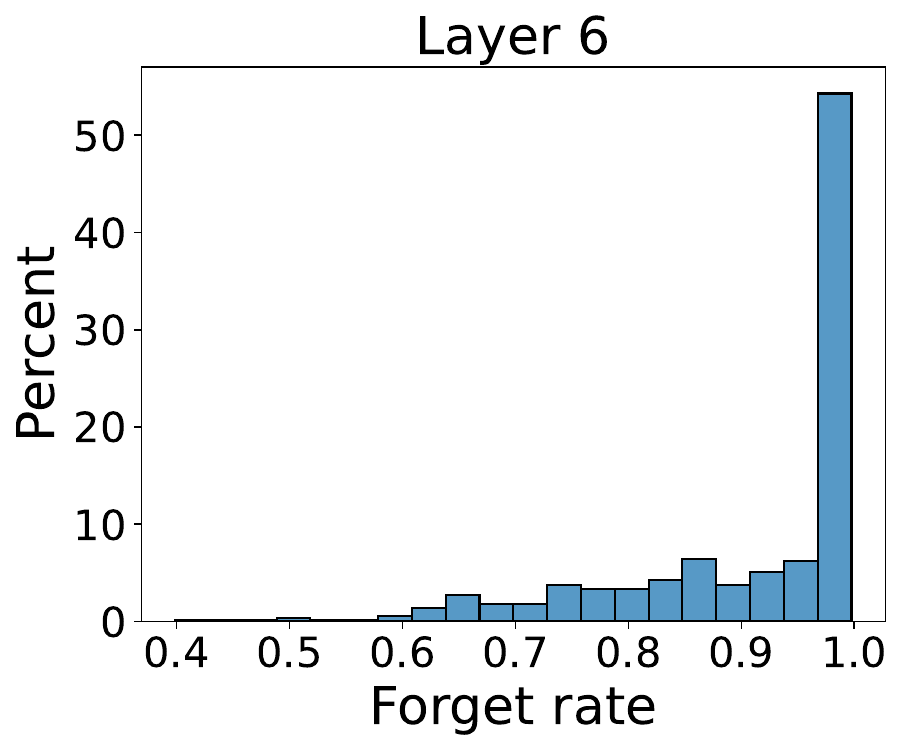}         
     \end{subfigure}
     \hfill
     \begin{subfigure}[b]{0.3\textwidth}
         \centering
         \includegraphics[width=\textwidth]
         {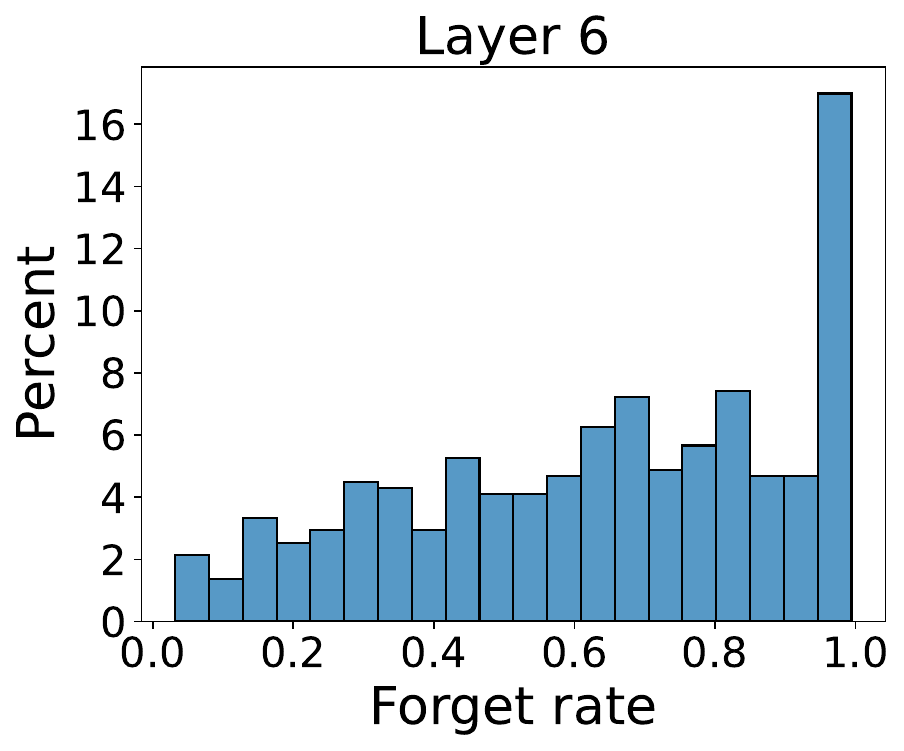}
     \end{subfigure} 
      \hfill
     \begin{subfigure}[b]{0.3\textwidth}
         \centering
         \includegraphics[width=\textwidth]
         {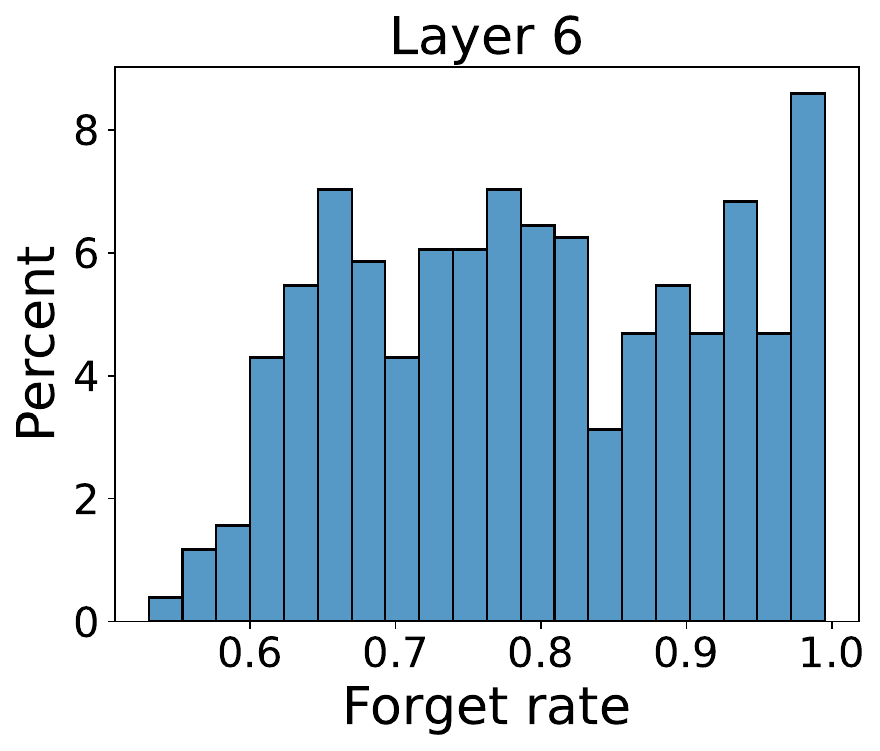}
     \end{subfigure}
     \newline
\end{figure}

\subsection{Configurations}
We list detailed hyper-parameters of our experiments here.

\label{token-mixing}
\begin{figure}
\label{fig:token-mixing}
     \centering
     \caption{Visualization of token mixing matrix in each layer.}
     \begin{subfigure}[b]{0.3\textwidth}
         \centering
         \caption{Layer 1}
         \includegraphics[width=\textwidth]{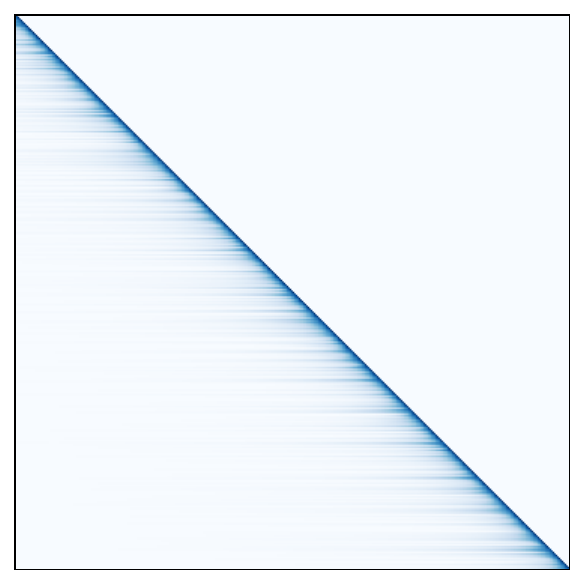}
         
     \end{subfigure}
     \hfill
     \begin{subfigure}[b]{0.3\textwidth}
         \centering
         \caption{Layer 1}
         \includegraphics[width=\textwidth]
         {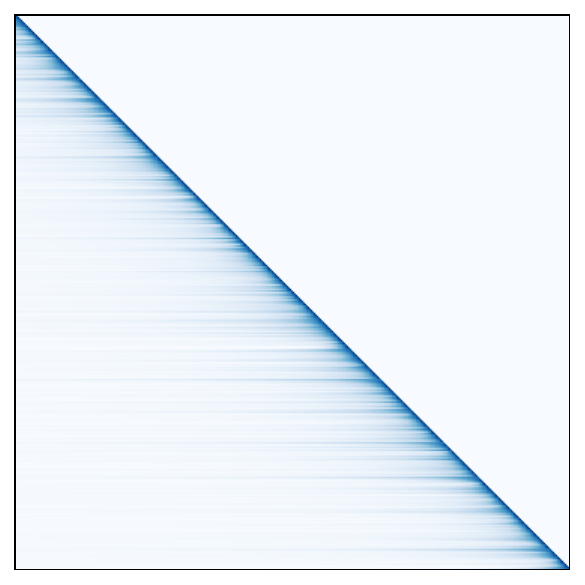}
     \end{subfigure} 
      \hfill
     \begin{subfigure}[b]{0.3\textwidth}
         \centering
         \caption{Layer 2}
         \includegraphics[width=\textwidth]
         {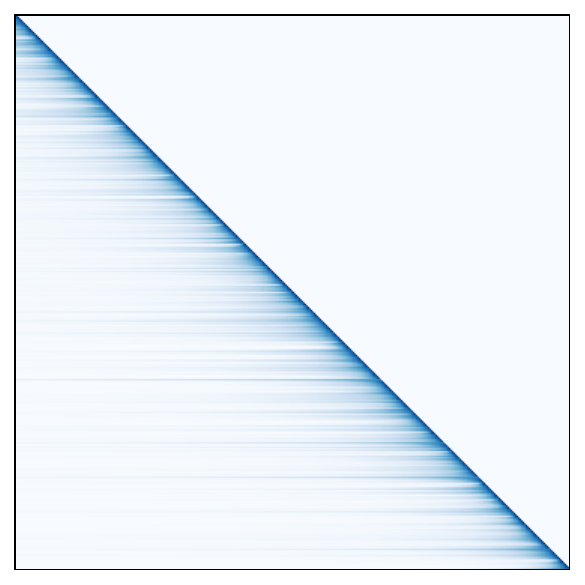}
     \end{subfigure}
     \newline

     \begin{subfigure}[b]{0.3\textwidth}
         \centering
         \caption{Layer 4}
         \includegraphics[width=\textwidth]{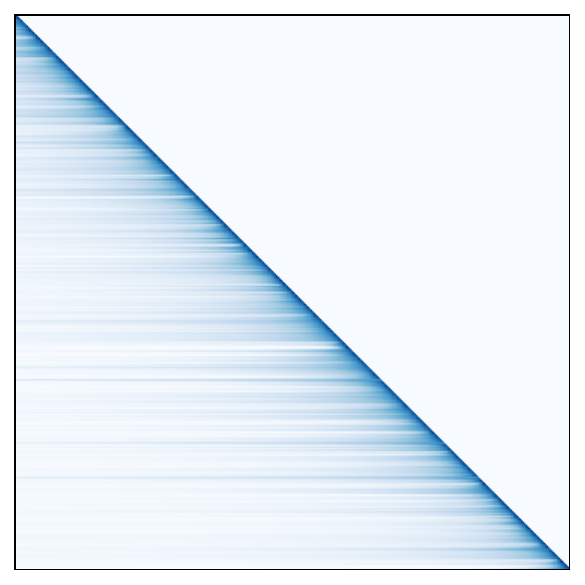}
         
     \end{subfigure}
     \hfill
     \begin{subfigure}[b]{0.3\textwidth}
         \centering
         \caption{Layer 5}
         \includegraphics[width=\textwidth]
         {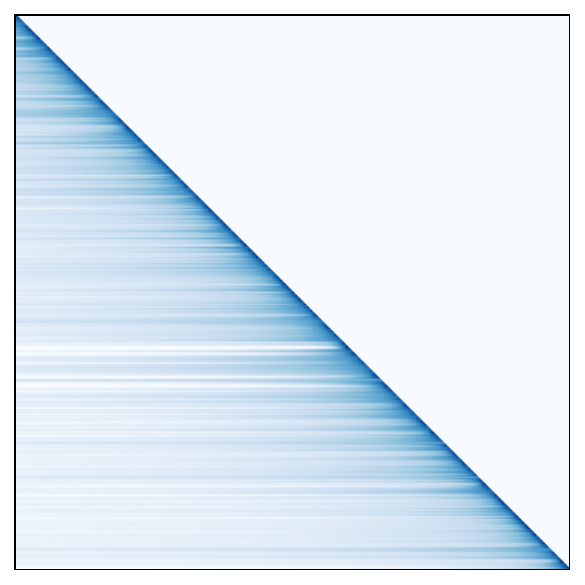}
     \end{subfigure} 
      \hfill
     \begin{subfigure}[b]{0.3\textwidth}
         \centering
         \caption{Layer 6}
         \includegraphics[width=\textwidth]
         {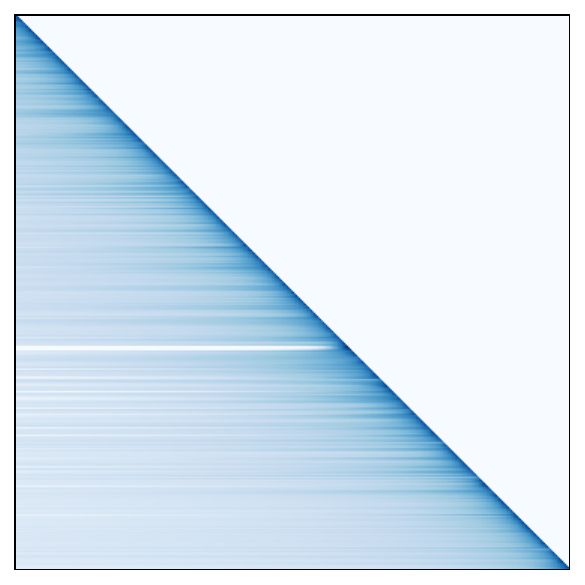}
     \end{subfigure}
     \newline
\end{figure}

\begin{table*}[!ht]
\small
\center
\setlength{\tabcolsep}{0.6cm}
{
\caption{Detailed training configurations used in our experiments. ``Total batch size'' means $\mathrm{batch\_per\_gpu} \times \mathrm{update\_freq} \times \mathrm{num\_gpus}$. ``ALM'' stands for Autoregressive Language Model. ``IM'' stands for Image Modeling.}
\label{configuration}
\begin{tabular}{l|l|l}
\hline\hline
 & AML  & IM \\
\hline\hline
Data                                              & WikiText-103    &ImageNet-1k                      \\
Tokenizer method & BPE    & - \\
Src Vocab size & 50265   & -\\
Sequence length    & 512   & -         \\
Total batch size & 128   & 2048                                            \\
Number of updates/epochs                            & 50k updates   & 300 epochs   \\
Warmup steps/epochs    & 4k steps    & 5 epochs      \\
Peak learning rate   & 5e-4     &2.5e-4                          \\
Learning rate scheduler  & Inverse sqrt    &cosine                   \\
Optimizer   & Adam  &Adamw                                     \\
Adam $\epsilon$   & 1e-8   & 1e-8                            \\
Adam $(\beta_1,\beta_2)$                          & (0.9, 0.98)   & (0.9, 0.98)            \\
Weight decay       & 0.2 & 0.1 \\                                       
Gradient clipping                                            &  -  & 1.0                                             \\
\hline\hline
\end{tabular}}

\end{table*}

\end{document}